\documentclass[letterpaper, 10 pt, conference]{ieeeconf}  % Comment this line out if you need a4paper

\IEEEoverridecommandlockouts                              % This command is only needed if 
\usepackage{amsmath} % assumes amsmath package installed
\usepackage{amssymb}  % assumes amsmath package installed
\usepackage{float}
\usepackage{cleveref}
\usepackage{multirow}
\usepackage{bbm}
\usepackage{textcomp}
\usepackage{xcolor}
\usepackage{subcaption}
\usepackage{caption}
\usepackage{graphicx}
\usepackage{amsmath,amssymb,amsfonts}
\usepackage{algorithmic}
\usepackage{mathtools}
\usepackage[linesnumbered,ruled,vlined]{algorithm2e}
\usepackage{nccmath} %%%%%%% for ceqn
\usepackage{dsfont}
\usepackage{bbm}
\usepackage{multicol}
\usepackage{cite}
\usepackage{xcolor}
\usepackage{amsmath}
\usepackage{hhline}

\SetKwRepeat{Do}{do}{while}%
\SetKwInput{KwInput}{Input}                % Set the Input
\SetKwInput{KwOutput}{Output}    
\DeclareMathOperator{\E}{\mathbb{E}}
\usepackage{tikz}
\usepackage{lipsum}

\newcommand\copyrighttext{%
  \footnotesize \textcopyright 2022 IEEE. Personal use of this material is permitted.  Permission from IEEE must be obtained for all other uses, in any current or future media, including reprinting/republishing this material for advertising or promotional purposes, creating new collective works, for resale or redistribution to servers or lists, or reuse of any copyrighted component of this work in other works.}
\newcommand\copyrightnotice{%
\begin{tikzpicture}[remember picture,overlay]
\node[anchor=south,yshift=10pt] at (current page.south) {\fbox{\parbox{\dimexpr\textwidth-\fboxsep-\fboxrule\relax}{\copyrighttext}}};
\end{tikzpicture}%
}

\title{\LARGE \bf
Unsupervised Reinforcement Learning for Transferable Manipulation Skill Discovery
}

\author{Daesol Cho$^{1}$, Jigang Kim$^{1}$ and H. Jin Kim$^{1*}$
\thanks{$^{1}$Department of Mechanical and Aerospace Engineering, Seoul National University, Seoul, Korea}
}

\begin{document}

\maketitle
\thispagestyle{empty}
\pagestyle{empty}

\copyrightnotice
% \lipsum[1-10]

\begin{abstract}
Current reinforcement learning (RL) in robotics often experiences difficulty in generalizing to new downstream tasks due to the innate task-specific training paradigm. To alleviate it, unsupervised RL, a framework that pre-trains the agent in a task-agnostic manner without access to the task-specific reward, leverages active exploration for distilling diverse experience into essential skills or reusable knowledge. For exploiting such benefits also in robotic manipulation, we propose an unsupervised method for transferable manipulation skill discovery that ties structured exploration toward interacting behavior and transferable skill learning. It not only enables the agent to learn interaction behavior, the key aspect of the robotic manipulation learning, without access to the environment reward, but also to generalize to arbitrary downstream manipulation tasks with the learned task-agnostic skills. Through comparative experiments, we show that our approach achieves the most diverse interacting behavior and significantly improves sample efficiency in downstream tasks including the extension to multi-object, multitask problems.

\end{abstract}

\section{INTRODUCTION}

While deep reinforcement learning has shown considerable progress toward solving complex robotic control tasks in the presence of extrinsic rewards \cite{gu2017deep}\cite{yamada2020motion}, these advances produced agents that are unable to generalize to new downstream tasks beyond the one they were trained to solve. Humans, on the other hand, are able to acquire skills with minimal supervision and apply them to solve a variety of downstream tasks. Inspired by human's flexible capability, recently, unsupervised RL, a framework that trains the agent without access to the environment reward supervision, has emerged as a promising paradigm for developing RL agent that can generalize to new downstream tasks.

In the unsupervised RL setting, the agent is first pre-trained in a downstream-task-agnostic environment without access to the environment reward supervision, and then transferred to the various tasks with the environment rewards. As the goal of unsupervised pre-training is to have data-efficient adaptation for some downstream tasks, some prior works \cite{sharma2020emergent, eysenbach2018diversity} address the pre-training problem by leveraging the skill learning method. However, such skill-based techniques are only applicable within the variation of the objectives in the \emph{same} environment, and these cannot be transferred into the task variation in totally different environments due to the limited flexibility of the learned skills.

More fundamental challenge is that in the context of the unsupervised RL, skill learning is intimately connected to the efficient exploration of the given environment and vice-versa. That is, the skills discovered by the agent depend on the regions of the state space covered during exploration. If exploration is ineffective, the learned skills cannot properly characterize meaningful parts of the state space in the environment, degrading the performance for downstream tasks. Even though some exploration methods are proposed for obtaining diverse data distribution \cite{pathak2017curiosity, burda2018exploration, bellemare2016unifying}, most of these methods are still limited to exploration based on the novelty of the proprioceptive states rather than encouraging exploration towards the interaction behavior, which is a key aspect of the robotic manipulation skill learning. Conversely, efficient exploration cannot be performed without proper skills since no practical exploration method would be able to exhaustively explore all possible states. Thus, we argue that the development of algorithms specialized to simultaneously address such proper skill learning and diverse, effective exploration for manipulation tasks is of paramount importance to benefit from unsupervised pre-training.

\begin{figure}[]
\centering
\includegraphics[width=\linewidth]{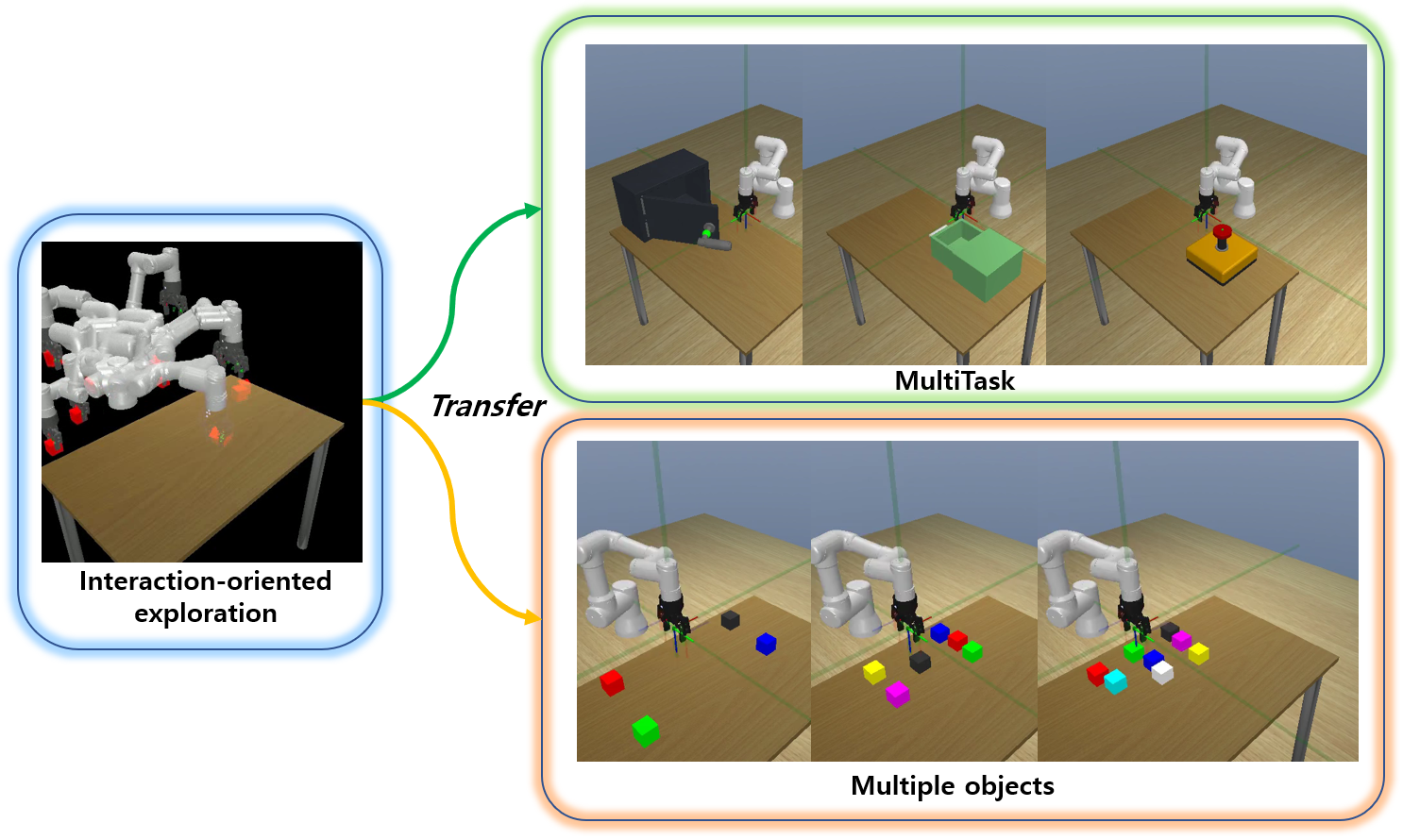}
\caption{Task-agnostic skills naturally emerged by interaction-oriented exploration can significantly improve the sample efficiency in learning diverse downstream manipulation tasks.}

\label{fig:thumbnail}
\vspace{-0.5cm}
\end{figure}

In this work, we address the challenges described above through a framework for the unsupervised training on robot manipulation that learns interactive behaviors without access to task-specific rewards. Specifically, we consider the mutual information-based two-phase approach with specialized policy network design, which enables addressing the diverse-exploration challenge, but also distilling the generated experience in the form of reusable task-agnostic skills. In the first phase, the agent explores the environment without prior knowledge of downstream tasks, while learning task-agnostic, interactive skills through intrinsic rewards based on mutual information. Then in the second phase, given the skills trained in the first phase, the agent is transferred to solve various downstream tasks with as few environment interactions as possible. Such a setup evaluates the agent's ability to adapt to new tasks after pre-trained only once in an unsupervised manner.

To summarize, our work makes the following contributions.
\begin{itemize}
    \item To the best of the author's knowledge, this is the first work on an unsupervised method for transferable manipulation skill discovery, which generates task-agnostic skills for transferring to a wide range of robot manipulation tasks.
    \item The proposed method achieves the most diverse, interactive exploration for unsupervised pre-training compared to other baselines.
    \item After pre-training only once with the proposed method, 2-5 times improvements in sample efficient training have been achieved in various downstream tasks including the extension to multi-object, multitask problems.
\end{itemize}

\section{Related Works}

A fundamental problem in RL is exploring the state space, especially in cases where the reward is sparse. For tackling this problem, task-agnostic approaches or exploiting various inductive biases that correlate positively with structured, efficient exploration are proposed. Prior works include state-visitation counts \cite{bellemare2016unifying}\cite{ostrovski2017count}, eigendecomposition-based exploration \cite{machado2017laplacian}, curiosity/similarity-driven exploration \cite{pathak2017curiosity}\cite{warde2018unsupervised}, prediction model's uncertainty-based exploration \cite{burda2018exploration}\cite{pathak2019self}, mutual information-based exploration \cite{eysenbach2018diversity}\cite{sharma2019dynamics}\cite{zhao2020mutual}. In contrast, our work, aiming at robot manipulation learning, is different in the aspect of drawing motivation from interaction-oriented exploration and distilling the experience into reusable skills rather than just trying to discover unseen, new states.

To enable sample efficient RL for various downstream tasks, several researchers have taken inspiration from skill transfer or multiple-task learning. They have shown that utilizing different modules across different tasks reduces the interference on each module \cite{devin2017learning}\cite{qureshi2019composing}. But pre-defined skills or sub-policies need human engineering for effectiveness in specific tasks, while our work enables the agent to learn and transfer skills in a task-agnostic way. A different set of approaches are learning the single integrated policy by addressing the entire tasks all at once with a carefully designed network \cite{yu2020meta}\cite{yang2020multi}, or improving the optimization technique \cite{yu2020gradient}\cite{suteu2019regularizing}. Instead of distilling the multitaskable knowledge into a policy by experiencing all tasks, our approach could be considered as learning necessary backbone features for downstream tasks, which allows a significant margin in data efficiency and flexibility of the learned skills.

\section{Preliminary} \label{preliminary}
% MDP & RL formulation, 
In this study, we consider a latent augmented Markov Decision Process (MDP) defined by a state space $\mathcal{S}$, action space $\mathcal{A}$, discount factor $\mathcal{\gamma}$, reward function $R$, transition probability $p(s'|s,a)$, and latent space $\mathcal{Z}$. The objective is to obtain a policy $\pi(a|s,z)$ to maximize the expected sum of rewards $E[\sum\limits^{T}_{t=0}R(s_t, a_t, z_t)]$, where the latent variable $z$ is sampled from some distribution $p(z)$, and states are sampled according to initial state distribution $\rho(s_0)$, $a_t \sim \pi(a_t|s_t,z_t)$, $s_{t+1} \sim p(s_{t+1}|s_t,a_t)$. All of the latent $z$ in these definitions could be replaced with the goal $g$ in goal space $\mathcal{G}$ when we consider the goal-conditioned MDP.

% Mutual Information 
One of the important quantities in the following training process is mutual information (MI) $I(X;Y) = \int_{\mathcal{X}} \int_{\mathcal{Y}}p(x,y)\log\frac{p(x,y)}{p(x)p(y)} \,dx\,dy$, which can be equivalently expressed as $\mathcal{H}(X) - \mathcal{H}(X|Y)$, where $\mathcal{H}$ is entropy. It is a measure of the mutual dependence between the two variables $X, Y$. If $X, Y$ is independent, $I(X;Y)$ is 0, and otherwise, it has a positive value. This mutual information could be used as an intrinsic reward to induce the interacting behavior of the robot.

% MCP structure
Another quantity is a compositional policy network structure, called multiplicative compositional policies (MCP) \cite{peng2019mcp}. It represents the policy by multiplication of some primitives and gating network, and it is expressed as follows,

\begin{equation}\label{eqn:MCP_latent}
    \pi(a|s,z) = \frac{1}{Z(s,z)}\prod_{i=1}^{N} \pi_i(a|s)^{W_i(s,z)},\:\:\:\: W_i(s,z)\ge0
\end{equation}
where $\pi_i$ is the $i$th primitive, and $W_i$ is a positive gating weight for $\pi_i$, and $Z$ is a partition function that ensures the output action distribution is normalized. In this study, we use $N=8$ and the latent variable $z$ is augmented to the MCP, and it will be explained in section \ref{subsec:Method-A-play}. With the assumption of Gaussian distribution $\pi_i$, the output of the MCP could be represented by Gaussian distribution as follows,

\begin{align}\label{eqn:MCP_primitive_latent}
\begin{split}
    {}& \mu^j(s,z) = \frac{1}{\sum\limits_{l=1}^N\frac{W_l(s,z)}{\sigma^j_l(s,z)}}\sum\limits_{i=1}^N\frac{W_i(s,z)}{\sigma_i^j(s,z)}\mu_i^j(s,z) \\
    {}&\quad \sigma^j(s,z) = \left(\sum\limits_{i=1}^N\frac{W_i(s,z)}{\sigma_i^j(s,z)}\right)^{-1}
\end{split}
\end{align}
where $\mu^j_i$, $\sigma^j_i$ is the $j$th element of the $\pi_i$'s mean and variance. Unlike other compositional policy architecture such as an additive policy which consists of the weighted sum of primitive policies, MCP could be effective when the effect of each primitive's action should be in a non-zero-sum way. That is, simultaneous activation of each primitive could be more effective and could represent more complex expression and concurrent behavior.

\section{Method} \label{method}

\subsection{How to induce interaction-oriented exploration}\label{subsec:Method-A-play}

To learn interaction-oriented exploration behaviors without any external task reward, some intrinsic reward that implicitly induces interacting behavior is needed. One of the possible approaches is utilizing mutual information (MI) $I(S_o;S_r)$ as an intrinsic reward, where $S_o$ is the object's position, $S_r$ is the robot's gripper position. Then, by definition of $I(S_o;S_r) = \mathcal{H}(S_o) - \mathcal{H}(S_o|S_r)$, the policy maximizes the uncertainty of the object's state while minimizing the uncertainty of the object's state when given the gripper's state. By using this MI as a reward, the robot is intrinsically motivated to interact with an object. That is, the robot should grasp an object and place it in other states here and there.

As MI between the two variables from unknown distribution is generally not tractable, we propose to use Jensen-Shannon Divergence (JSD)-based mutual information estimation \cite{nowozin2016f}, which is expressed as follows,
% JSD detailed derivation
\begin{equation}\label{eqn:JSD}
\begin{split}
  &I(S_o;S_r)=D_{JSD}(P||Q),  (P : P_{S_o S_r}, Q : P_{S_o}P_{S_r}) \\
  &= \int_{\mathcal{X}}q(x)\sup_{t\in dom_{f^*}}\Big\{t\frac{p(x)}{q(x)}-f^*(t)\Big\}\,dx \\
  &\ge \sup_{T\in\mathcal{T}}\Big(\int_{\mathcal{X}}p(x)T(x)\,dx - \int_{\mathcal{X}}q(x)f^*(T(x))\,dx  \Big)\\
  &= \sup_{\psi_1 \in \Psi}\E_{x \sim P}[T_{\psi_1}(x)]-\E_{x \sim Q}[f^*(T_{\psi_1}(x))] \\
  &= I_{\psi_1}(S_o;S_r) \\
\end{split}
\end{equation}

where the distribution $P$ is a joint distribution between $S_o$ and $S_r$, $Q$ is a product of marginal distribution of each $S_o$ and $S_r$. The first and second equalities hold from the definition of $f$-divergence \cite{nowozin2016f}, and inequality holds from the Jensen's inequality when swapping the integration and supremum operations. Following the \cite{nowozin2016f}, we substituted the $T(x) = log(2)-log(1+e^{-g_{\psi_1}(x)})$, where $g_{\psi_1}$ is a neural network parameterized by $\psi_1$, and $f^*(x) = -log(2-e^x)$. By maximizing the lower bound of \eqref{eqn:JSD}, where the expectation is over the data collected by the policy, we could train the MI estimator $I_{\psi_1}$.

However, we empirically found that the estimated $I(S_o,S_r)$ alone makes some interactive behavior of the agent, but does not induce the most diverse behavior. It could be due to the exploitation of the approximated lower bound rather than the true MI value. To incentivize more diverse behaviors when interacting with an object, we propose to use a additional diversity-driven component, referred from DADS \cite{sharma2019dynamics}. It introduces an information-theoretic objective as follows,

\begin{equation}\label{eqn:DADS}
% \nonumber
\begin{split}
% \int_{\mathcal{S}}\int_{\mathcal{Z}}\int_{\mathcal{S'}}
  &I(s';z|s)\\
  &=\iiint_{\mathcal{S,Z,S'}} p(s)p(s',z|s)\left[\log\frac{p(s',z|s)}{p(s'|s)p(z|s)}\right] \,ds'\,dz\,ds\\
  &=\E_{z,s,s'\sim p(z,s,s')}\left[log\frac{p(s'|s,z)}{p(s'|s)}\right] \\
  &=\E_{z,s,s'\sim p(z,s,s')}\left[log\frac{q_{\psi_2}(s'|s,z)}{p(s'|s)}\right] \\
  &+\E_{s,z\sim p(s,z)}\left[D_{KL}(p(s'|s,z)||q_{\psi_2}(s'|s,z))\right] \\
  &\ge \E_{z,s,s'\sim p(z,s,s')}\left[log\frac{q_{\psi_2}(s'|s,z)}{p(s'|s)}\right] = I_{\psi_2}(s';z|s)\\
\end{split}
\end{equation}

where the first equality holds by the definition of conditional mutual information, and the inequality holds by non-negativity of KL-divergence. $I(s';z|s)$ can be decomposed into $\mathcal{H}(s'|s) - \mathcal{H}(s'|s,z)$, and it encourages the agent to make the next state $s'$ as diverse as possible given the current state $s$, while reducing the uncertainty of $s'$ when given $s$ and latent $z$ sampled from the prior uniform distribution $p(z)$. So $z$ could mean `skill' that discriminates state distribution followed by some policy that maximizes the reward $I(s';z|s)$. That is, each different $z$ induces each different trajectory while maintaining low variance of trajectory when the same $z$ is given, and $q_{\psi_2}$ is called `skill dynamics' due to this property. The output distribution of the $q_{\psi_2}$ is modeled as a mixture of Gaussian distributions with the parameterized neural network, and the MI estimator $I_{\psi_2}$ could be trained by maximizing the lower bound of \eqref{eqn:DADS}, where the expectation is over the data collected by the policy.

By substituting the state $s$ with $S_o$, the two MI estimations could be combined together as an intrinsic reward,
\begin{equation}\label{eqn:total_reward}
    \hat{r} = I_{\psi_1}(S_o;S_r) + I_{\psi_2}(S_o';z|S_o)
\end{equation}
and by maximizing \eqref{eqn:total_reward} through RL, we could expect that the robot is naturally induced to perform interaction-oriented exploration, while learning skills. All of these intrinsic rewards are formulated in an unsupervised way, and there is no external task reward signal for manipulation behavior. As the reward is represented by MI that contains the latent $z$, the overall MDP is augmented with $z$, and the policy is represented as in \eqref{eqn:MCP_latent}.

% Second step in training process
\subsection{Transfer to goal-conditioned reinforcement learning}\label{subsec:Method-B-transfer}

\begin{figure}[t]
  \includegraphics[width=\linewidth, height=5.5cm]{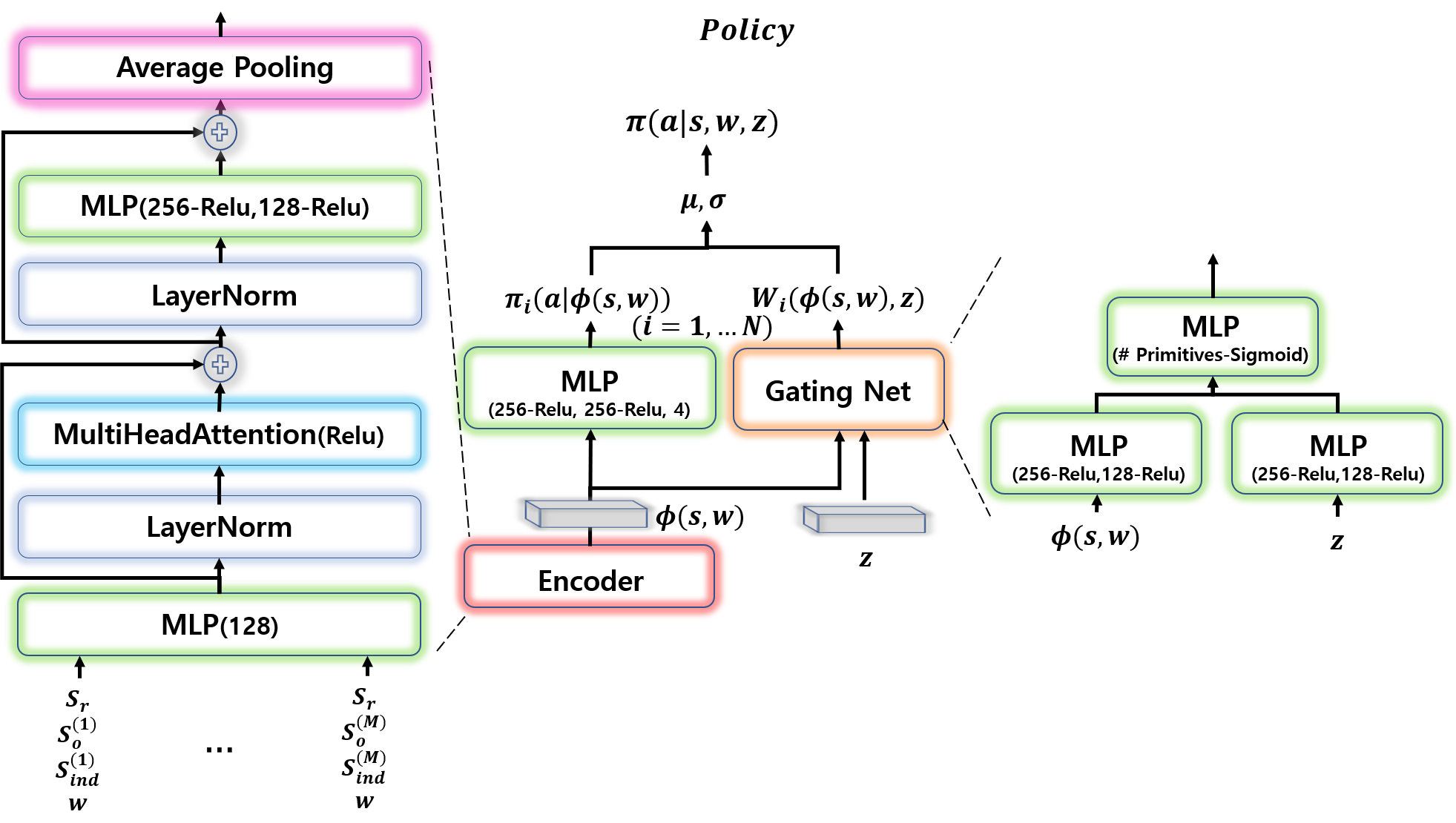}
  \caption{Policy architecture. In a single-object setting, $\phi(s,w)$ is replaced with $s$, and the encoder and $w$ is not used. When transferring to GCRL, the latent $z$ on the policy is replaced with the goal $g$.}
  \label{fig:architectures}
  \vspace{-0.5cm}
\end{figure}
Through the unsupervised pre-training phase in \ref{subsec:Method-A-play} , the agent is encouraged to learn interaction-oriented exploration behavior, and it leads the policy $\pi(a|s,z)$ to be trained to have skill-agnostic primitives $\pi_i$ and gating network $W_i$ that regularizes the activation of each primitive. Then, we can say that the primitives $\pi_i$ implicitly contain the necessary information about interaction; which action distribution is most frequently used when interacting with an object, and which combination of action distribution is enough to represent the possible action outputs that cover the state space where the interaction behavior occurred. That is, the primitives $\pi_i$ can be a backbone of the policy for any kind of task that needs interaction with something.

% Considering these points,
% As the primitives $\pi_i$ contain interacting behavior implicitly
For deploying such properties of the learned primitives $\pi_i$, we can transfer them to the goal-conditioned reinforcement learning (GCRL) by retraining only the gating network while replacing the latent $z$ in \eqref{eqn:MCP_latent} with goal $g$, and freezing the weights of the $\pi_i$ to maintain the primitives as they are. Then, the policy can be expressed as follows.

\begin{equation}
    \pi(a|s,g) = \frac{1}{Z(s,g)}\prod_{i=1}^{N} \pi_i(a|s)^{W_i(s,g)},\:\:\:\: W_i(s,g)\ge0
    \label{eqn:MCP_goal}
\end{equation}
which can be used for any type of GCRL problems.

\subsection{Extension to multiple objects}\label{subsec:Method-C-MultiObject}

\begin{algorithm}[t]
\caption{Learning task-agnostic skills and transferring to GCRL.}
\label{algo:ALGO1} 
\DontPrintSemicolon
    % $Q$, 
    NOTE: Exclude the $w$ in single object setting. \\
    Initialize parameters of $I_{\psi_1}, I_{\psi_2}$, $\pi$, buffer $\mathcal{B}$. \\
    \While{not converged}{
        \For{$i=1, 2, \dots, K$}{
            Get $\{(s_t, z_i, w_i, a_t, r_t)\}_{t=0}^{n}\sim\pi(a|s,w_i,z_i)$. \\
            Save $\{(s_t, z_i, w_i, a_t, r_t)\}_{t=0}^{n}$ in buffer $\mathcal{B}$. \\
            }
        Update $\psi_1, \psi_2$ by maximizing the \eqref{eqn:JSD}, \eqref{eqn:DADS}. \\
        Replace the reward $r$ with \eqref{eqn:total_reward}. \\
        Train the agent with SAC \cite{haarnoja2018soft}. \\ %(e.g. SAC).
    }
    % $Q$,
    Initialize buffer $\mathcal{B}$, $\pi(a|s,w,g)$ with fixed weight $\pi_i$ from $\pi(a|s,w,z)$ for transferring to GCRL.\\
    \While{not converged}{
        \For{$j=1, 2, \dots, K$}{
            Get $\{(s_t, g_j, w_j, a_t, r_t)\}_{t=0}^{n}\sim\pi(a|s,w_j,g_j)$. \\
            Save $\{(s_t, g_j, w_j, a_t, r_t)\}_{t=0}^{n}$ in buffer $\mathcal{B}$. \\
        }
        Train the agent with SAC \cite{haarnoja2018soft}. %(e.g. SAC+HER).
    }
\end{algorithm}
% \vspace{-0.25cm}

To extend the idea into multi-object setting, the agent should know how to deal with the variable number of objects and how to focus on a specific object despite the existence of the distractor objects. Thus, the state $s$ in \eqref{eqn:MCP_latent} should be represented with a fixed-dimension encoded feature $\phi(s,w)$ regardless of the number of objects, where $w$ is the one-hot encoded intention vector that indicates which object is the agent's interest. The dimension of $w$ is same as the maximum number of objects. The agent is rewarded when it achieves some desired behavior with respect to the object that corresponds to $w$. All of these properties could be achieved by using a transformer \cite{vaswani2017attention} based encoder, which has essentially the same utility as the graph neural network, because it has an attention mechanism that allows reusing the network while maintaining the scalability on the number of inputs.

% Transformer Policy 
Specifically, the state inputs in the $w$ augmented policy $\pi(a|s,w,z)$ consist of $s_{r}, s^{(i)}_o, s^{(i)}_{ind}$, where $s_{r}$ is the robot's states such as gripper's position, velocity, $s^{(i)}_o$ is the $i$th object's states such as object's position, velocity, $s^{(i)}_{ind}$ is one-hot encoded object indicator for distinguishing each different object. Without the indicator, the encoder cannot tell which state is from which object. For each $i$, $s_{r}, s^{(i)}_o, s^{(i)}_{ind}, w$ are concatenated, and these are passed into the modified transformer encoder (Fig \ref{fig:architectures}).

The modified parts in the encoder compared to the vanilla transformer encoder 
\cite{vaswani2017attention} are as follows: 1) Position embedding for distinguishing the spatial information among the inputs is not used as we care about an object's identity, not an object's spatial information in attention network structure. 2) The location of the layer normalization is changed right before the attention module and final mlp layer \cite{parisotto2020stabilizing} rather than followed by the add layer \cite{vaswani2017attention} for training stability. 3) Average pooling is applied in the encoder's outputs to obtain feature $\phi(s,w)$ for addressing the variable number of objects inputs. $\phi(s,w)$ is used instead of the state $s$ in primitives $\pi_i$ and gating weights $W_i$ in \eqref{eqn:MCP_latent}. That is, the policy can be represented as $\pi(a|s,w,z) = \frac{1}{Z(\phi(s,w),z)}\prod_{i=1}^{N} \pi_i(a|\phi(s,w))^{W_i(\phi(s,w),z)}$. Then, the same procedures in \ref{subsec:Method-A-play} and \ref{subsec:Method-B-transfer} except that $w$ is randomly sampled at the first timestep in each episode could be applied with the variable number of objects setting. The overall transfer process is summarized in Algorithm \ref{algo:ALGO1}.

\subsection{Extension to multitask learning}\label{subsec:Method-D-MultiTaskRL}

% \vspace{-0.25cm}
\begin{algorithm}[t]
\caption{MultiTask RL}
\label{algo:ALGO2} 
\DontPrintSemicolon
    Initialize parameters of $\pi$, $\alpha$, buffer $B_j$ for task $T_j$\\
    \While{not converged}{
        \For{$j=1, 2, \dots, K$}{
            Get $\{(s_t, T_j, g_j, a_t, r_t)\}_{t=0}^{t_{max}}\sim\pi(a|s,g_j,T_j)$. \\
            Save $\{(s_t, T_j, g_j, a_t, r_t)\}_{t=0}^{t_{max}}$ in buffer $\mathcal{B}_j$. \\
            }
        $D = (D_1, D_2, \dots, D_K) \sim (B_1, B_2, \dots, B_K)$
        Train the agent with any MultiTask RL algorithm with the sampled data $D$. \\
        }

\end{algorithm}
% \vspace{-0.5cm}

Given the learned primitives that implicitly represent the interacting behavior (\ref{subsec:Method-A-play}), we can leverage these properties to perform the multitask reinforcement learning (MTRL) efficiently. MTRL considers the standard RL algorithm (SAC \cite{haarnoja2018soft} in this work) except that it is conditioned on task distribution $p(T)$, where $T$ is in task space $\mathcal{T}$, and it could be expressed as follows,

\begin{equation}\label{eqn:MTRL}
    \max_{\pi}\E_{T_j \sim p(T)}\left[\E_{\tau_j \sim p^\pi(\tau_j|T_j)}\left[\sum^{t_{max}}_{t=0}r(s,a)+\alpha\mathcal{H(\pi)} \right] \right]
\end{equation}

where $T_j$ is one-hot embedding vector for the $j$th task, and $\tau_j$ is trajectory sampled from $T_j$. By optimizing the \eqref{eqn:MTRL}, the policy $\pi$ is trained to perform well on multiple tasks. We can use the same transfer process like in \ref{subsec:Method-A-play} and \ref{subsec:Method-B-transfer} except that the gating network's inputs are augmented with the task embedding $T$ when transferred to MTRL. That is, the policy is represented as $\pi(a|s,g,T) = \frac{1}{Z(s,g,T)}\prod_{i=1}^{N} \pi_i(a|s)^{W_i(s,g,T)}$, where $\pi_i$ are learned primitives (\ref{subsec:Method-A-play}) with fixed weights. MTRL process is summarized in Algorithm \ref{algo:ALGO2}.

\section{Experiments} \label{experiment}

In this study, the Fetch environment in the openAI gym is used to compare the mutual-information-based intrinsic reward's effectiveness, and custom UR3 with end-effector position and gripper control (20Hz, 4-dimensional action) environment is used for evaluation of the GCRL in multi-object, multitask, real-world experiments.

\subsection{Analysis of the learned exploration behavior}\label{subsec:Experiment-A-analysis-exploration}
To verify the effectiveness of the proposed method in terms of the interaction-oriented exploration behavior, we compare each different combination of mutual information-based unsupervised RL methods, where each method has the following properties,
\begin{itemize}
    \item MISC \cite{zhao2020mutual} : It uses Donsker-Varadhan (DV) representation to estimate $I(S_o;S_r)$ and only uses it as an intrinsic reward for exploration. It is the first and only work that shows success in stably picking an object without any external task reward.
    \item DIAYN \cite{eysenbach2018diversity} : It encourages the agent to explore as diverse as possible by maximizing $I(s;z)$. It suggests that the skill $z$ should control which states the agent visits, thus the agent is rewarded by visiting each different state space according to each different skill.
    \item Proposed (JSD+DADS) : Described in \ref{subsec:Method-A-play}. 
\end{itemize}

\begin{figure}[t]
  \includegraphics[width=\linewidth,height=5.2cm]{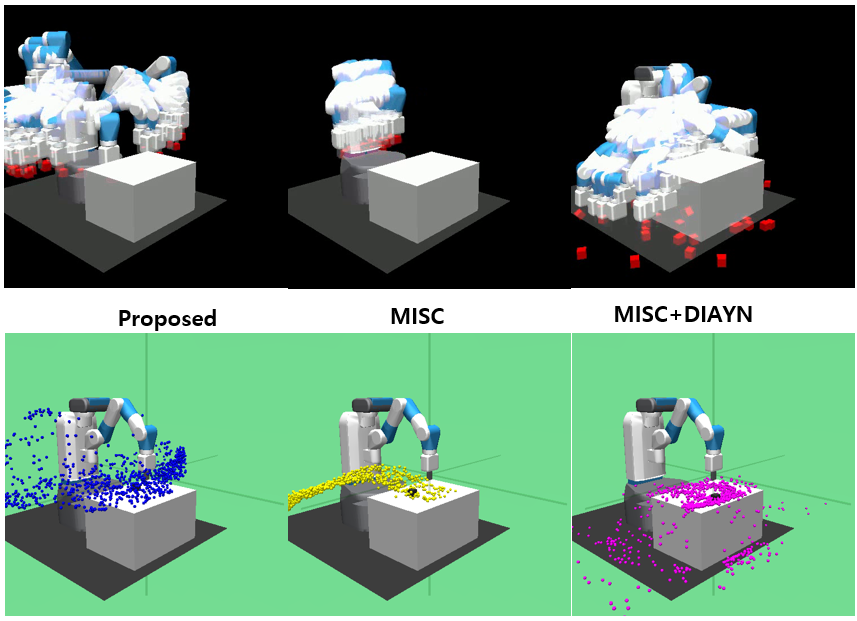}
  \caption{Visualization of the learned behavior (Top) and object's states (Bottom) for each different MI objective.}
  \label{fig:fetch diversity visualization}
  \vspace{-0.5cm}
\end{figure}

For comparison, 2000 trajectories (50 steps per trajectory) are gathered by $\pi(a|s,z)$ with multiple random seeds and latent $z$. Some of the collected trajectories and object's states are visualized in Fig \ref{fig:fetch diversity visualization}, and our proposed method shows more desirable interaction-oriented exploration behavior compared to other MI-based baselines.

\begin{figure}[H]
\centering
\includegraphics[width=0.6\linewidth,height=3.1cm]{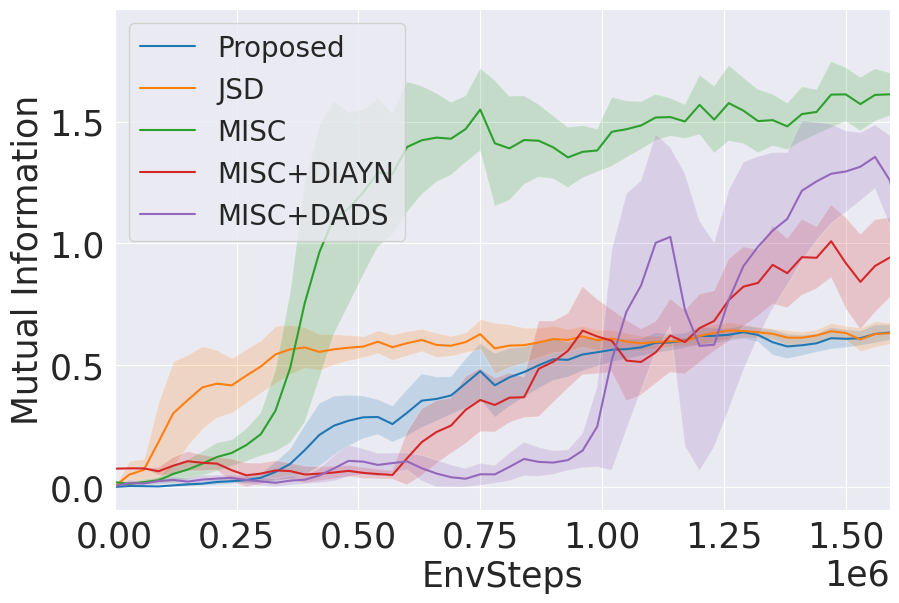}
\caption{Estimated value of $I(S_o;S_r)$. Higher values are not always good since the true MI values are unknown. Convergence rates and tendencies are more meaningful.}
\label{fig:estimated mi}
\vspace{-0.35cm}
\end{figure}

MISC uses Donsker-Varadhan (DV) representation to estimate $I(S_o;S_r)$, but this representation is known to be numerically unstable in estimating MI, which is verified on benchmark tests \cite{tsai2020neural}. Furthermore, it has a tendency to overestimate the MI (Fig \ref{fig:estimated mi}). As the agent exploits these properties, it collapses to almost the same behaviors once after grasping the object rather than diverse interaction behavior after grasping the object (Fig \ref{fig:fetch diversity visualization}).

Even though adding DIAYN is proposed in \cite{zhao2020mutual}, it frequently leads to missing the object. It is due to the different rewarding mechanisms for diversity. DIAYN rewards the occupation of different state distributions according to the intrinsic reward proportional to $\log q(z|s)$, where $q(z|s)$ is skill discriminator. Thus, even though the object is not moving in a specific region, the agent can receive the reward for diversity (e.g. pushing away the object out of the table). This property of DIAYN limits the utility of learned interaction primitives. However, DADS rewards according to the predictability of the next state (equation \eqref{eqn:DADS}), which encourages the agent to keep moving with the object. Thus, the agent does not receive the reward for diversity if the object is not moved.

To analyze the visualized results numerically, we compare each MI objective with 2 criteria; 1) the number of grasping among the states in trajectories for comparing the interaction ratio, 2) entropy of the object's states for comparing diversity. The number of grasping is computed by counting the number of states where the distance between the gripper and object is within a threshold (5 cm). The entropy is computed approximately by discretizing the 3D space into 3D bins and treating the ratio of the number of states in each bin as a probability. The values normalized by the total number of states and maximum entropy (uniform distribution) are shown in Fig \ref{fig:Primitive Analysis}-a.

As expected by visualization results, the proposed one shows a higher grasping ratio and object's state entropy compared to other baselines. While DIAYN brings higher entropy when added to MISC, the robot frequently pushes the object away rather than grasping (Fig \ref{fig:fetch diversity visualization}). DADS helps to increase the entropy while maintaining grasping because it optimizes the policy not only for diversity but also for predictability of the skill dynamics $q_{\psi_2}$. Furthermore, using Jensen Shannon Divergence (JSD) instead of DV representation (MISC) shows higher entropy and less failure in grasping due to the more stable, conservative estimation as the JSD-based estimation's value and variance are bounded \cite{tsai2021self}. These properties could be indirectly supported by looking into Fig \ref{fig:estimated mi}.

\vspace{-0.25cm}

\subsection{Analysis of the learned primitives}\label{subsec:Experiment-B-analysis-primitive}
Another factor that we look into is the distribution of each primitive. Collapse to similar primitives is not a desirable feature for transfer because the primitives not only should represent the backbone feature of interactive motions but also should be diverse to cover the possible action distributions that may be needed in the transferred downstream tasks. This could be explored through the local approximation to the perturbation sensitivity of $\pi$ by using the approximated Fisher Information.

While the Fisher matrix is typically computed with respect to the model parameters, we compute the modified diagonal Fisher $\hat{F}$ of the policy $\pi$, which is motivated by \cite{rusu2016progressive}, with respect to other representation $h^{(i)}$ such as gating weights $W_i$ and primitive's mean $\mu_i$. We define the diagonal matrix $\hat{F}$ with diagonal elements $\hat{F}(m,m)$, and derive the Average Fisher Sensitivity (AFS) of feature $m$ in the $i$th primitive as:

\begin{equation}
    \resizebox{\linewidth}{!}{%
    $\hat{F}^{(i)} = \E_{\rho^{\pi}(s,a)}\left[\frac{\partial log\pi}{\partial h^{(i)}} \frac{\partial log\pi}{\partial h^{(i)}}^T \right], AFS(i,m) =  \frac{\hat{F}^{(i)}(m,m) }{\sum_i \hat{F}^{(i)}(m,m)}$
    }
\end{equation}
where the expectation is over the joint state-action distribution $\rho^{\pi}(s,a)$ induced by the policy $\pi$. In practice, it is often useful to consider the AFS score per primitive $AFS(i) = \sum_m AFS(i,m)$, i.e. summing over all features in the $i$th primitive. AFS thus estimates how much the policy relies on each primitive to compute the output.

\begin{figure}[t]%
\centering
\begin{subfigure}{0.25\textwidth}
\centering
\includegraphics[width=\linewidth, height=2.8cm]{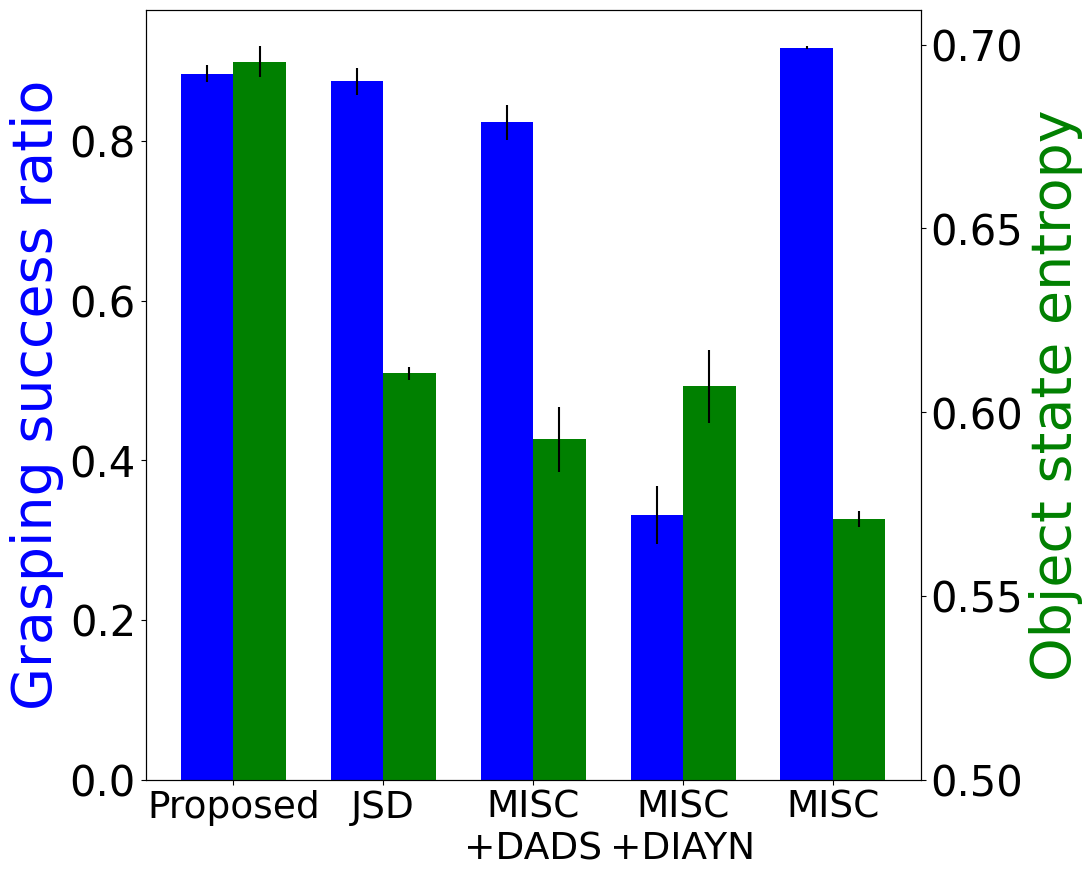}%
\vspace{-0.2cm}
\caption{Grasping ratio \& $\mathcal{H}$(Object)}%
\label{Primitive Analysis:subfig:a}%
\end{subfigure}%\vfill%
\medskip
\begin{subfigure}{0.25\textwidth}
\centering
\includegraphics[width=\linewidth,height=2.8cm]{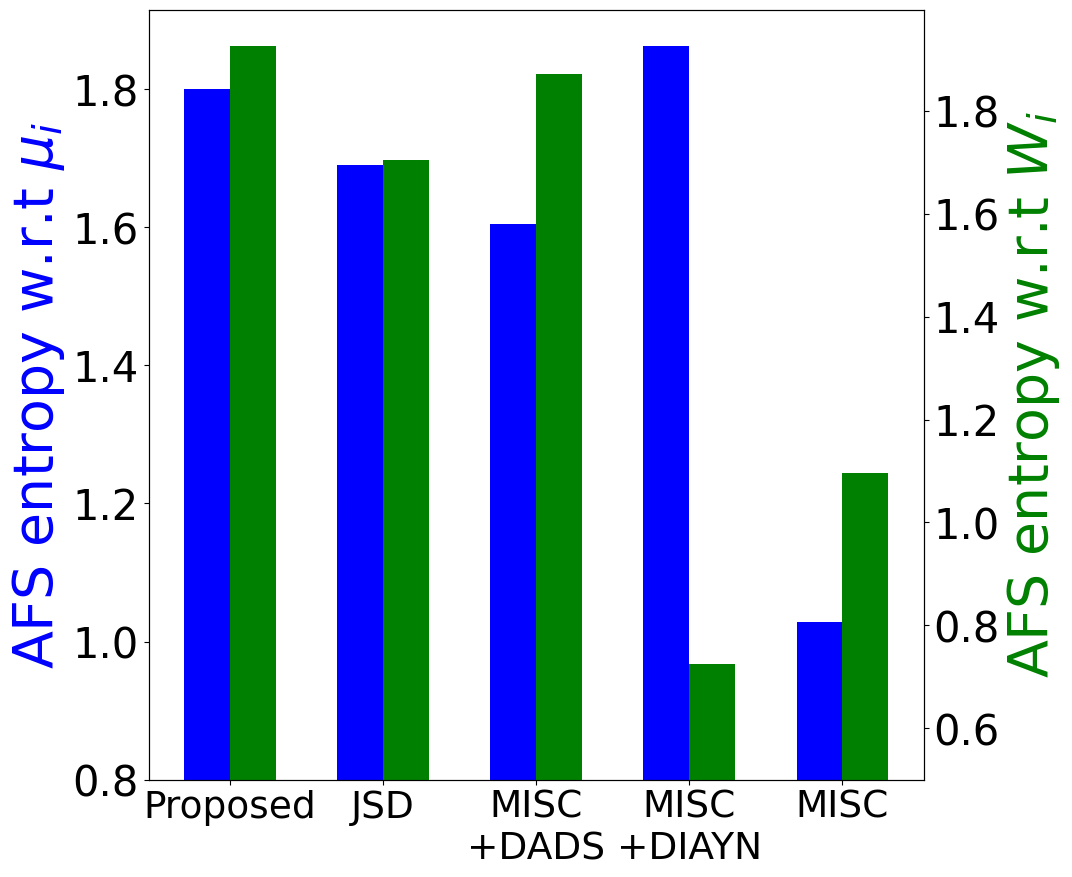}%
\vspace{-0.2cm}
\caption{AFS entropy}%
\label{Primitive Analysis:subfig:b}%as
\end{subfigure}%\hfill%
\medskip
\vspace{-0.3cm}
\begin{subfigure}{0.25\textwidth}
\centering
\includegraphics[width=\linewidth, height=2.4cm]{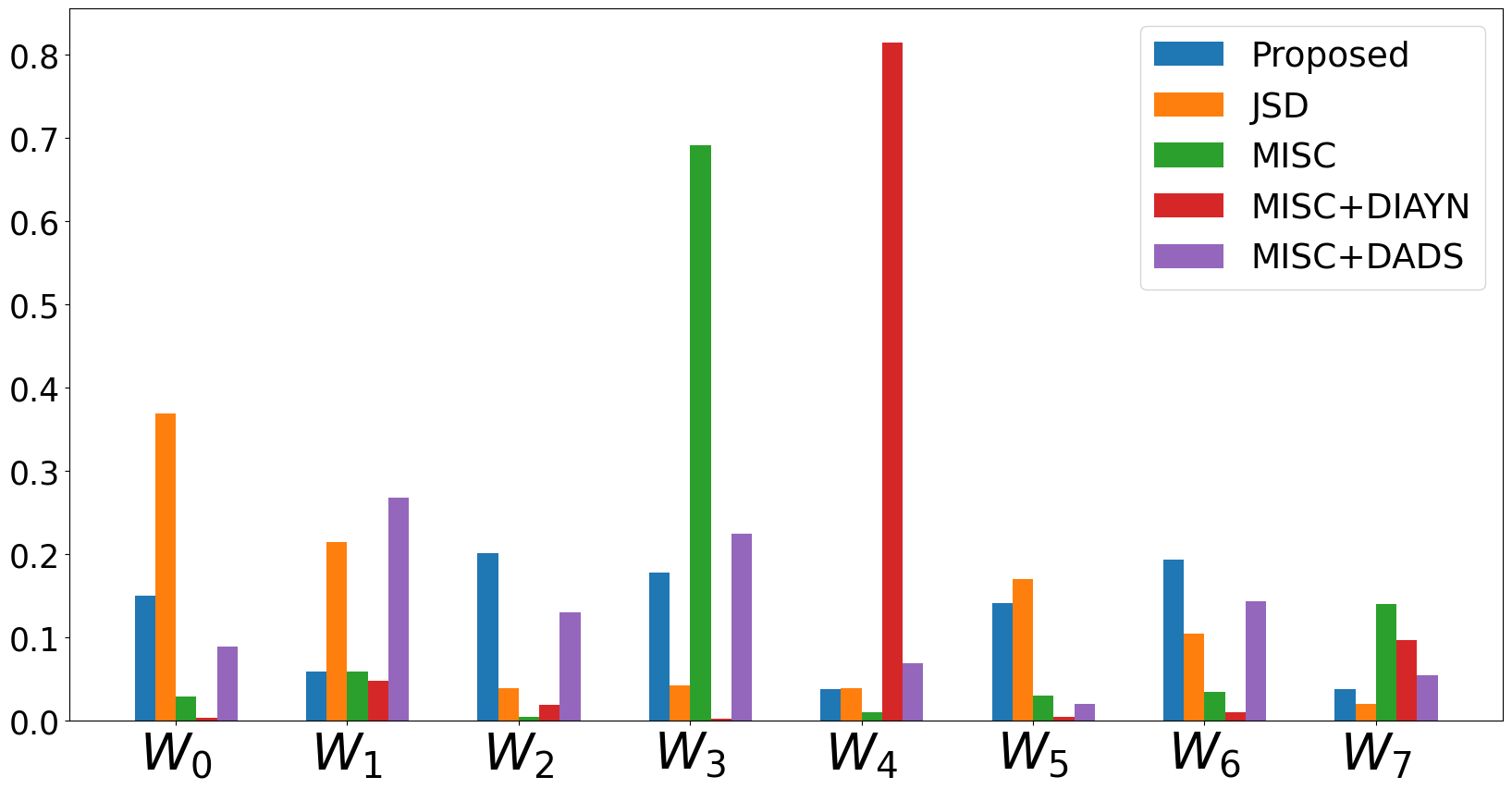}%
\vspace{-0.2cm}
\caption{AFS w.r.t gating weights $W_{i}$}%
\label{Primitive Analysis:subfig:c}%
\end{subfigure}%
\medskip
\begin{subfigure}{0.25\textwidth}
\centering
\includegraphics[width=\linewidth, height=2.4cm]{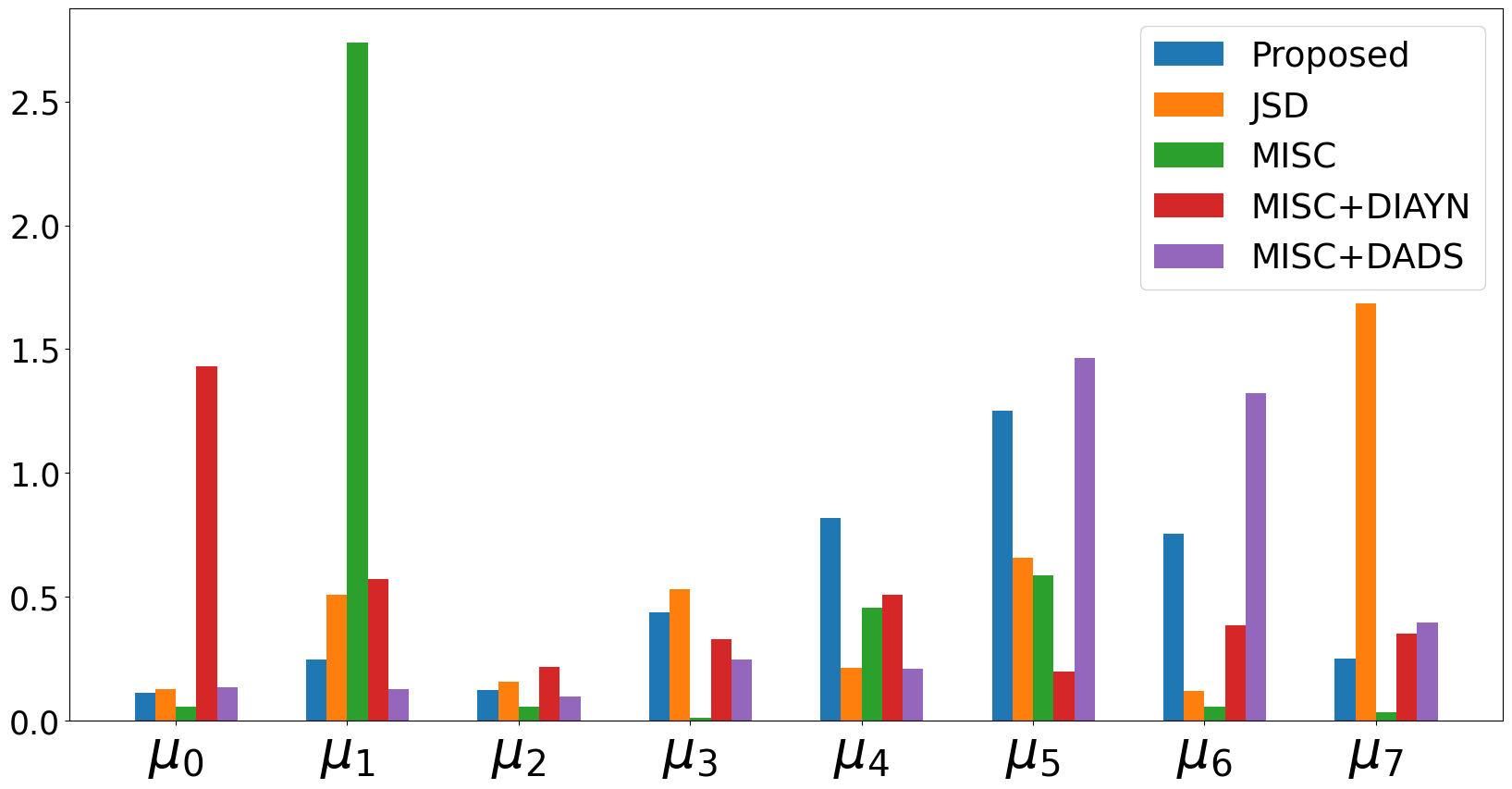}%
\vspace{-0.2cm}
\caption{AFS w.r.t primitive mean $\mu_i$}%
\label{Primitive Analysis:subfig:d}%
\end{subfigure}%
\vspace{-0.3cm}
\caption{Analysis of the learned behavior and primitives.}\label{fig:Primitive Analysis}
\vspace{-0.65cm}
\end{figure}

As AFS is computed along with each primitive (Fig \ref{fig:Primitive Analysis}-c,d), we use the normalized value of AFS as a pseudo categorical probability (Just for comparison. It is not probability by definition) to compute the AFS' pseudo entropy for comparing the distribution of the sensitivity along with each primitive (Fig \ref{fig:Primitive Analysis}-b). The desirable feature of the primitives is low sensitivity (high entropy of AFS) as it means each primitive performs its own roles rather than collapsing to similar distributions with a few different ones (Further verified by t-sne visualization in Fig \ref{fig:Real world experiments}). If collapse occurs, the sensitivity of a few primitives will surge like in MISC case in Fig \ref{fig:Primitive Analysis}-c,d.

For minor ablation, there could be a question about whether the learned primitives $\pi_i$ really represent the interaction behavior implicitly. To validate it, 2000 trajectories are collected by the policy, $\pi(a|s) = \frac{1}{Z(s)}\prod_{i=1}^{N} \pi_i(a|s)^{W_i(s)}$, with the primitives $\pi_i$ trained by each MI objective, and randomly initialized gating network $W_i(s)$ (Fig \ref{ablation:a}). Then, the interaction ratio is computed by checking whether the number of states where the distance between object and gripper is within a threshold (5 cm) is more than 60\% in a trajectory. The proposed method shows the highest interaction ratio, and we can say that implicit representation of the interaction behavior in $\pi_i$ is a reasonable conjecture and it is helpful for interaction-oriented exploration in the transferred downstream tasks.

\begin{figure}[t]\label{fig:Trnasfer to GCRL}%
\centering
\begin{subfigure}{0.25\textwidth}
\centering
\includegraphics[width=\linewidth]{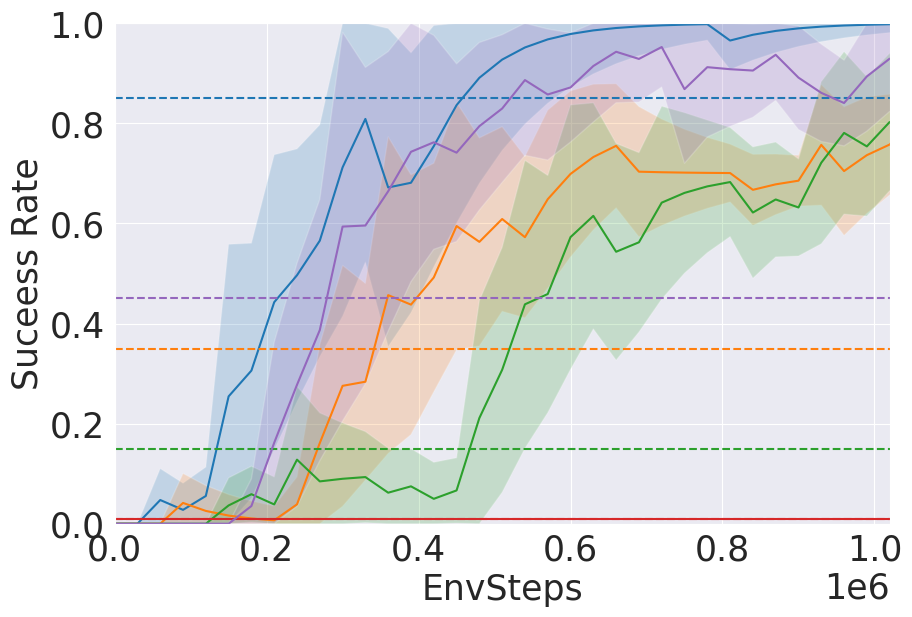}%
% \captionsetup{justification=centering}
%\vspace{-0.1cm}
\caption{Transfer to 1 object}%
\label{Transfer 1 obj:subfig:a}%
\end{subfigure}%\vfill%
\medskip
\begin{subfigure}{0.25\textwidth}
\centering
\includegraphics[width=\linewidth]{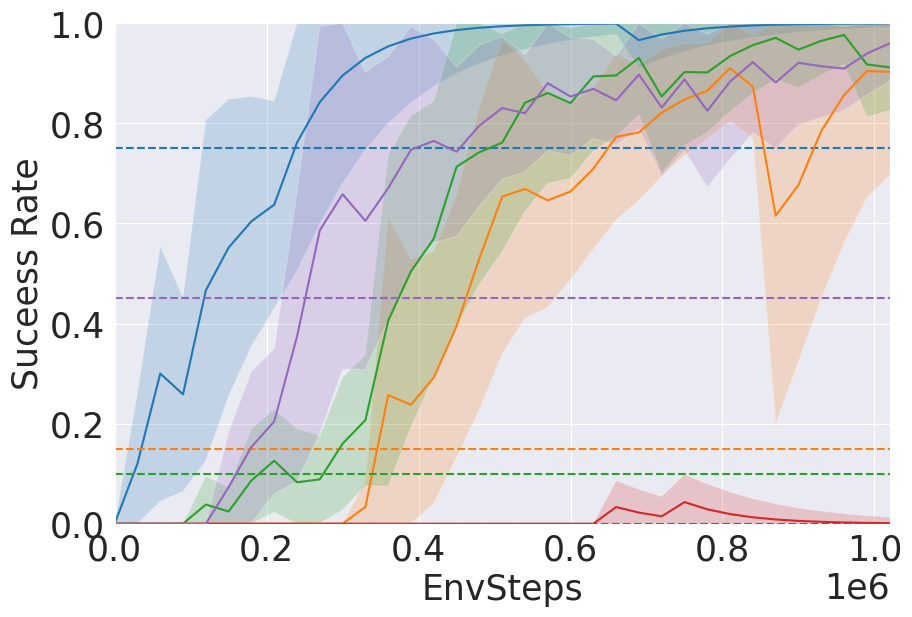}
%\vspace{-0.1cm}
\caption{Transfer to 4 objects}%
\label{Transfer 4 obj:subfig:b}%
\end{subfigure}%\hfill%
\medskip
\vspace{-0.3cm}
\begin{subfigure}{0.25\textwidth}
\centering
\includegraphics[width=\linewidth]{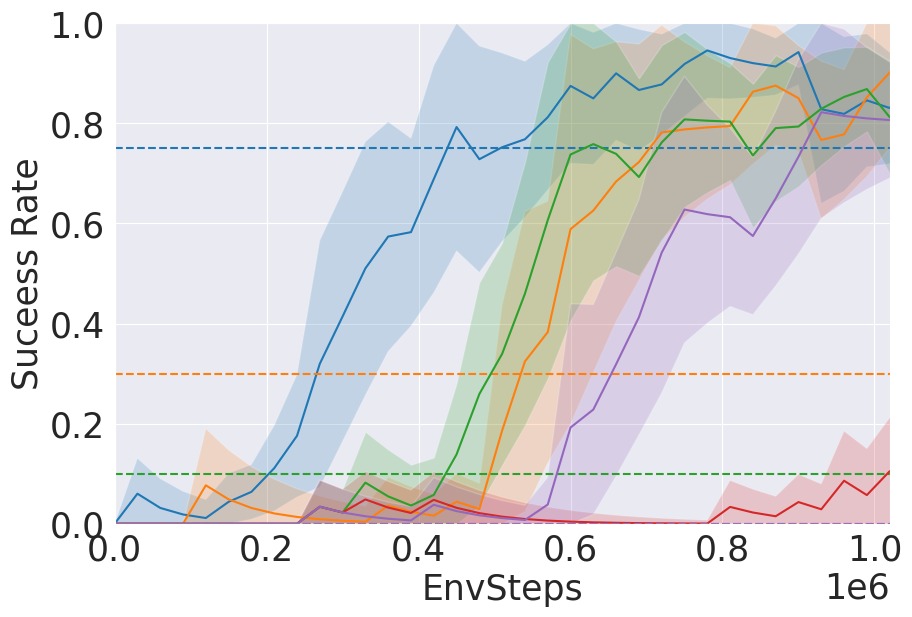}%
%\vspace{-0.1cm}
\caption{Transfer to 6 objects}%
\label{Transfer 6 obj:subfig:c}%
\end{subfigure}%
\medskip
\begin{subfigure}{0.25\textwidth}
\centering
\includegraphics[width=\linewidth]{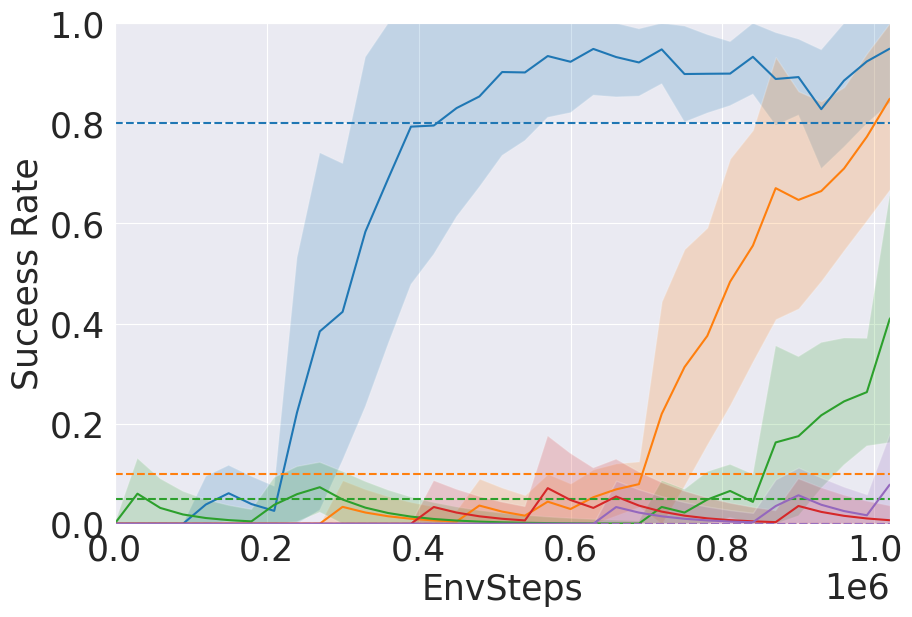}%
%\vspace{-0.1cm}
\caption{Transfer to 8 objects}%
\label{Transfer 8 obj:subfig:d}%
\end{subfigure}%
\vspace{-0.2cm}
\begin{subfigure}{0.5\textwidth}
\centering
\includegraphics[width=0.7\linewidth, height=0.7cm]{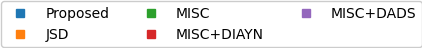}%
\end{subfigure}%
\vspace{-0.1cm}
\caption{The results of transferring to GCRL. The dashed line indicates real-world evaluation results of each method after 500k steps training (without finetuning in the real-world).}
\label{fig:Transfer to GCRL}
\vspace{-0.3cm}
\end{figure}

\subsection{Transfer to GCRL}\label{subsec:Experiment-B-comparison_multi_object}

To compare the effectiveness of the primitives learned by the proposed unsupervised pre-training method, we compare the UR3 pick\&place tasks with other MI-based unsupervised baselines and supervised RL baselines. Over the all experiments, a sparse reward that indicates whether the task is succeeded or not is used with the goal relabeling technique, HER \cite{andrychowicz2017hindsight}. In multi-object setting, the learned primitives are transferred to GCRL where the sparse reward is given only when the agent performs the given task with the object corresponding to $w$. As the attention based feature $\phi(s,w)$ is used instead of state $s$ in the multi-object setting, we could transfer the learned primitives $\pi_i$ without constraint on the number of objects. Thus, the learned primitives in the 4 objects setting are transferred to 6 \& 8-object setting.

Over the all different number of objects, the proposed method achieves the large sample efficiency margins in transferred downstream tasks, even with the multiple distractor objects in inputs. Most of the other mutual information-based baselines show slow convergence or failures within 1M steps (Fig \ref{fig:Transfer to GCRL}) due to the lack of primitive's expressivity stems from not enough exploration and skill discovery in the pre-training phase. % 

\begin{figure}[t]
\centering
\begin{subfigure}{0.22\textwidth}
\centering
\includegraphics[width=\linewidth, height = 2.9cm]{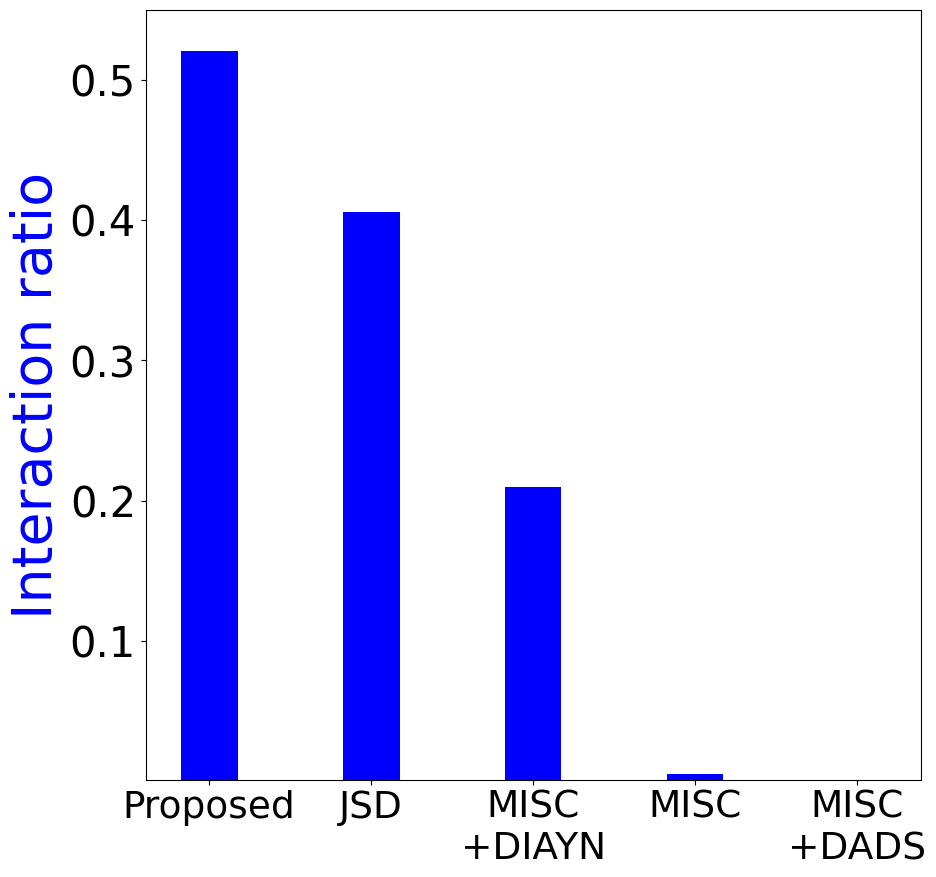}
% \vspace{-0.25cm}
\caption{Interaction ratio}%
\label{ablation:a}%
\end{subfigure}
\medskip
\begin{subfigure}{0.24\textwidth}
\centering
\includegraphics[width=\linewidth, height = 2.9cm]{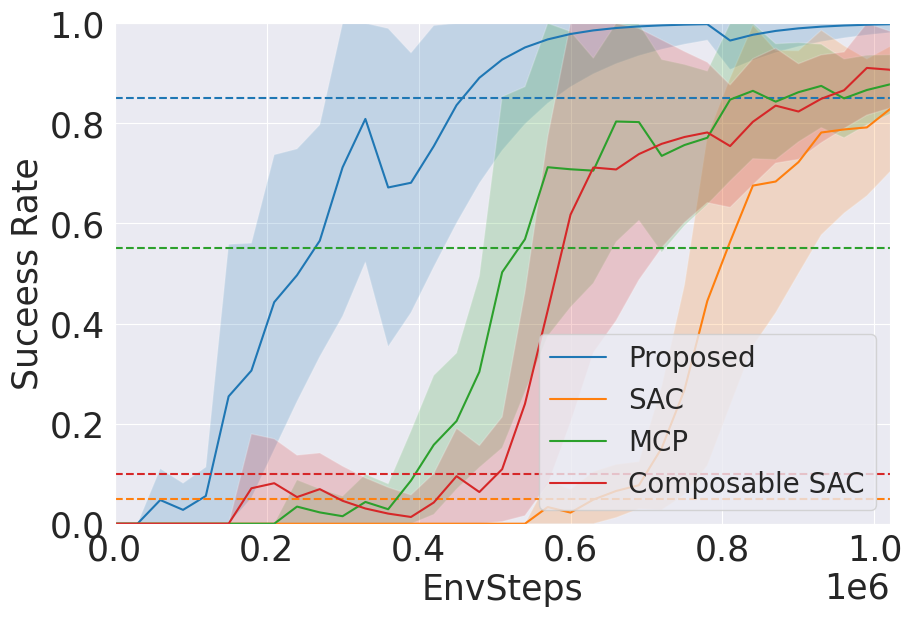}%
\caption{Comparison with others}%
\label{ablation:b}%
\end{subfigure}
\vspace{-0.2cm}
\caption{Ablation study : (a) Interaction ratio of the policy with the learned $\pi_i$ and randomly initialized $W_i$. (b) comparison with the supervised RL. The dashed line indicates real-world evaluation results (same as Fig \ref{fig:Transfer to GCRL}).} 
\label{fig:Ablation}
\vspace{-0.5cm}
\end{figure}

For ablation, in a single-object setting, we also compare with standard supervised RL baselines (Fig \ref{ablation:b}) to validate the advantage of the unsupervised pre-training. The proposed method shows a significant improvement in sample efficiency than SAC (vanilla RL with Gaussian policy), and MCP \cite{peng2019mcp} (SAC with the policy structure in \eqref{eqn:MCP_goal}), Composable SAC \cite{qureshi2019composing} (SAC with attention-based additive policy rather than multiplicative one). It supports the advantage of unsupervised pre-training that brings powerful exploration capability and extracts essential parts of the manipulation skills. We do not include the results on multi-object settings as the proposed method only succeeded.

% Real World Experiments
The trained policies are also evaluated in real-world pick\&place experiments with the UR3 (Fig \ref{fig:Transfer to GCRL}, \ref{ablation:b} with the dashed lines). Even though the success rates are slightly lower compared to simulation due to the sim-to-real gap such as dynamics, measurement error, the overall tendencies are maintained. Some of the failures of the policies that show some success in simulation are due to the safety constraint of the UR3 hardware, which limits the joint from moving any further when the inputs into the robot arm produce too much agile movement. We suspect that the primitives with higher AFS entropy effectively regularizes the action outputs as the low perturbation sensitivity means a slower change in action probability when inputs are changed, and it could lead to the less agile movement.

\begin{figure}[t]
  \includegraphics[width=\linewidth,height=6.5cm]{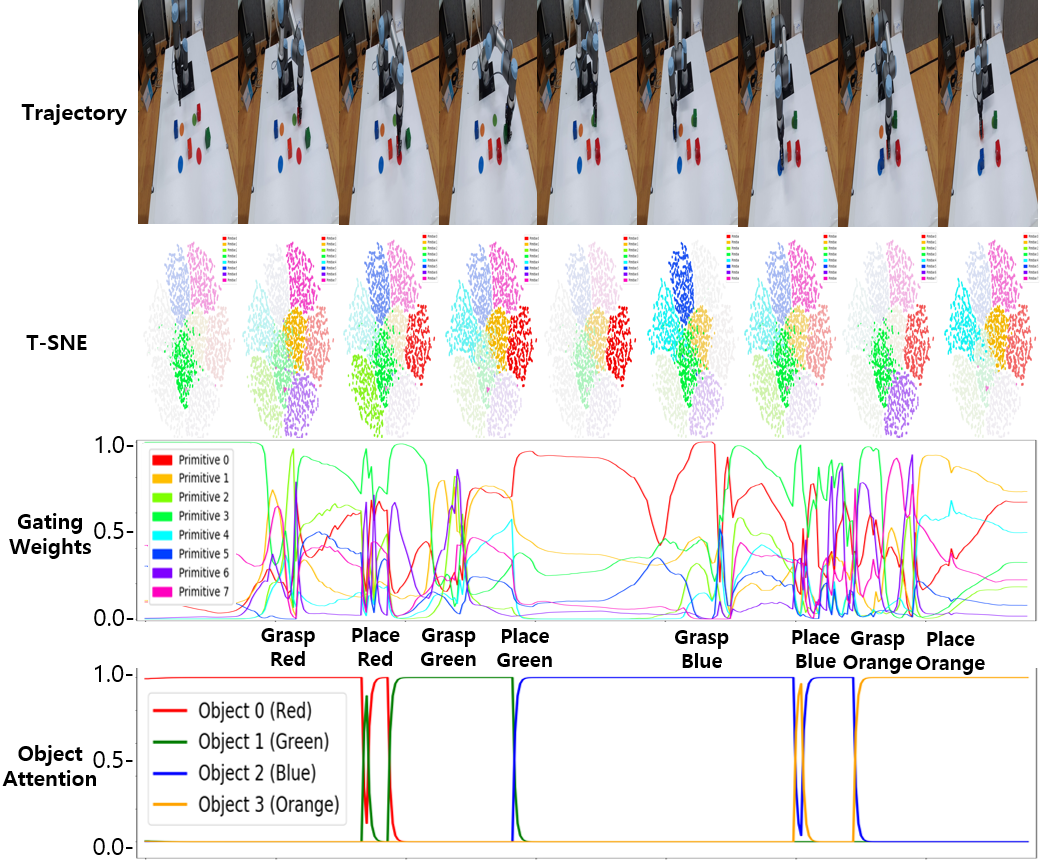}
  \caption{Real-world experiment results in a sequential task problem. Each primitive's gating weight and t-sne visualization of each primitive's output are represented with the same color, and the attention weight is plotted with the same color as each corresponding object (Red, Green, Blue, Orange). As $w$ is chosen by a simple scripted rule, there are some jitterings in attention weights when the agent slightly fails at placing the object to the right goal.} % (Red, Green, Blue, Orange)
  \label{fig:Real world experiments}
  \vspace{-0.5cm}
\end{figure}

% Simple application (sequential task)
To test whether the entire primitives are properly combined together rather than a few ones working across all timesteps, the trained policy can be used in solving sequential task problems like hierarchical reinforcement learning (HRL), where high-level policy's output corresponds to intention vector $w$. As our study is not about HRL, we just assume that $w$ is specified by a scripted rule at every step, and evaluate sequential tasks in real-world experiments (Fig \ref{fig:Real world experiments}). The robot is asked to arrange 4 objects into each goal state in the order of red, green, blue, orange. We could verify that every moment that the robot has to change its behavior, the gating weights are regularized to combine each primitive properly rather than collapsed to a few primitives. Also, the robot has the most attention to the object that corresponds to $w$ at that timestep. Further experimental results can be found in the supplementary video.

\subsection{Transfer to MTRL}\label{subsec:Experiment-C-comparison_MTRL}
% MTRL detail
We evaluate our method in the environments including pick\&place with/without obstacle, pushing, drawer \& door open/close, button press, where all tasks need interaction with the robot's gripper. These are chosen from the metaworld environment \cite{yu2020meta} developed for meta/multitask RL for robotics control. Building on top of these tasks, we further extend the tasks to be randomly generated goal-conditioned tasks, unlike fixed goals in the original metaworld, and modify the shaped reward into a sparse reward to make the training process more difficult. We also propose to learn the temperature $\alpha$ in SAC for adjusting the entropy of the policy on a per-task basis, i.e. using a parameterized model to represent $\alpha_j \sim f_\zeta(T_j)$. Without the learnable $\alpha$, the agent may stop exploring once all easier tasks are solved.

The proposed method is compared with other frequently referred baselines with SAC: 1) MT-SAC where task embedding $T$ is concatenated in policy \& critic. 2) MHMT-SAC built upon MT-SAC with independent heads for tasks. Both baselines are proposed in \cite{yu2020meta}. 3) PCGrad \cite{yu2020gradient} which uses similar architecture like in MT-SAC, but projects the conflicting task gradient into other task's normal plane to avoid a local optimum due to the gradient conflict. The conflict means negative cosine similarity between task gradients in PCGrad.

\begin{figure}[t]
\centering
\begin{subfigure}{0.25\textwidth}
\centering
\includegraphics[width=\linewidth, height=2.65cm]{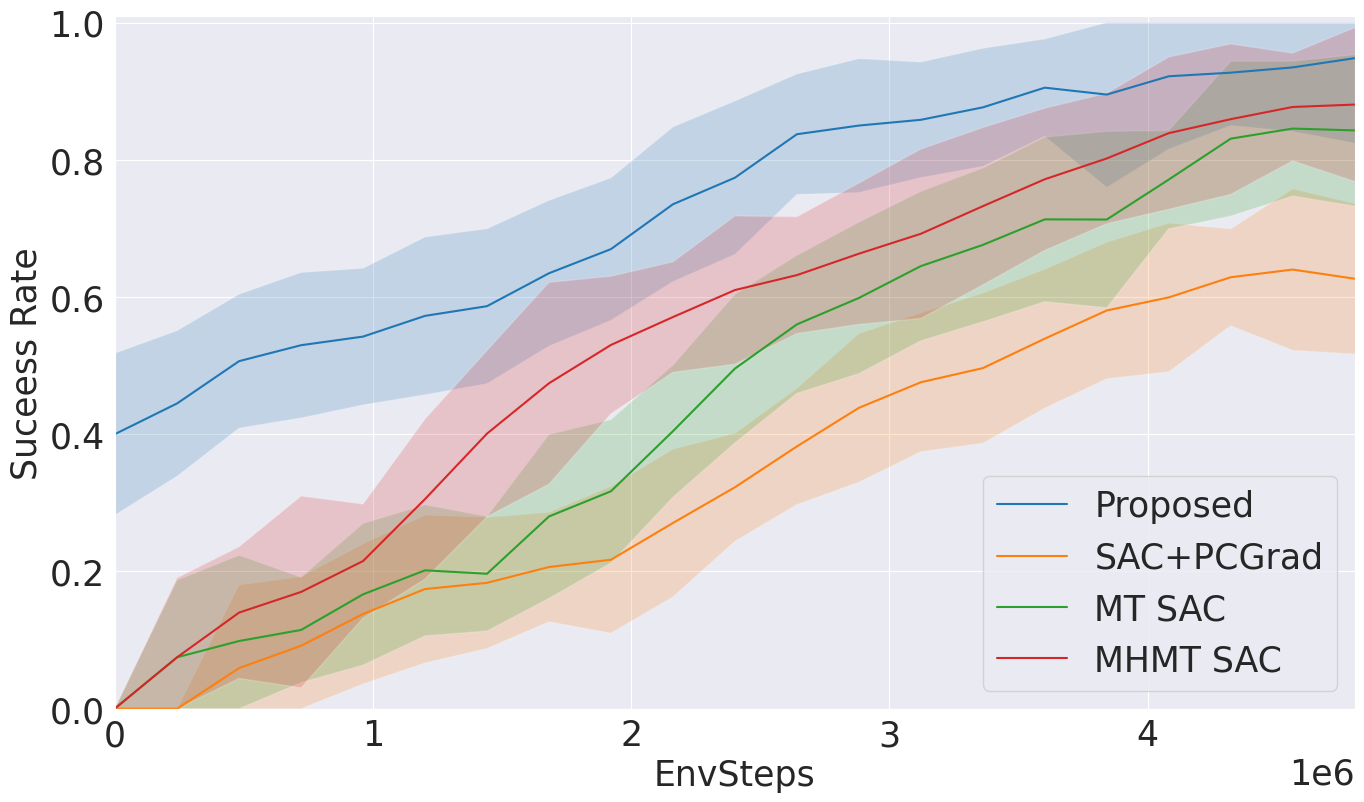}%
\vspace{-0.25cm}
\caption{All}%
\label{Transfer to MTRL All:subfig:a}%
\end{subfigure}%\vfill%
\medskip
\begin{subfigure}{0.25\textwidth}
\centering
\includegraphics[width=\linewidth, height = 2.65cm]{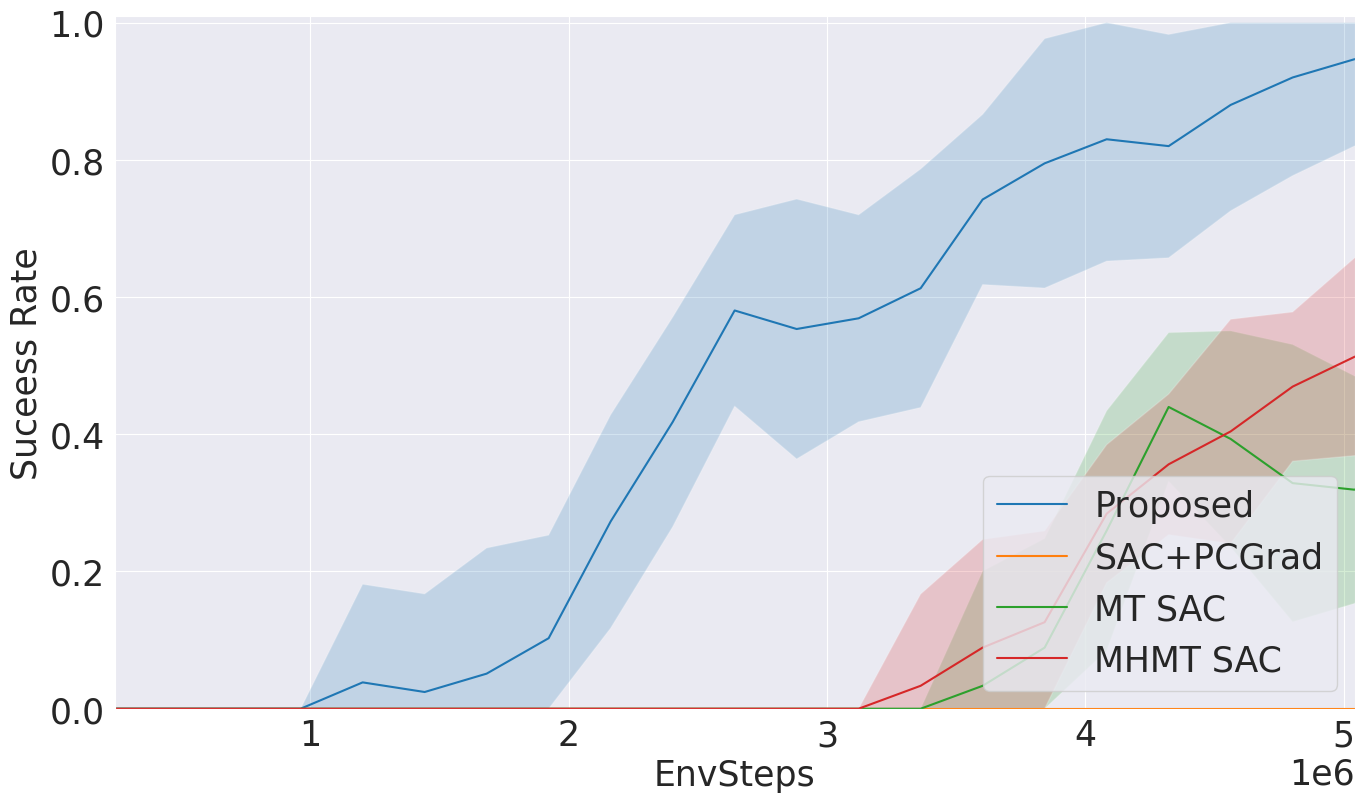}%
\vspace{-0.25cm}
\caption{Pick\&Place with obstacle}%
\label{Transfer to MTRL PickAndPlace:subfig:b}%
\end{subfigure}%\hfill%
\medskip
\vspace{-0.4cm}

\begin{subfigure}{0.25\textwidth}
\centering
\includegraphics[width=\linewidth, height=2.65cm]{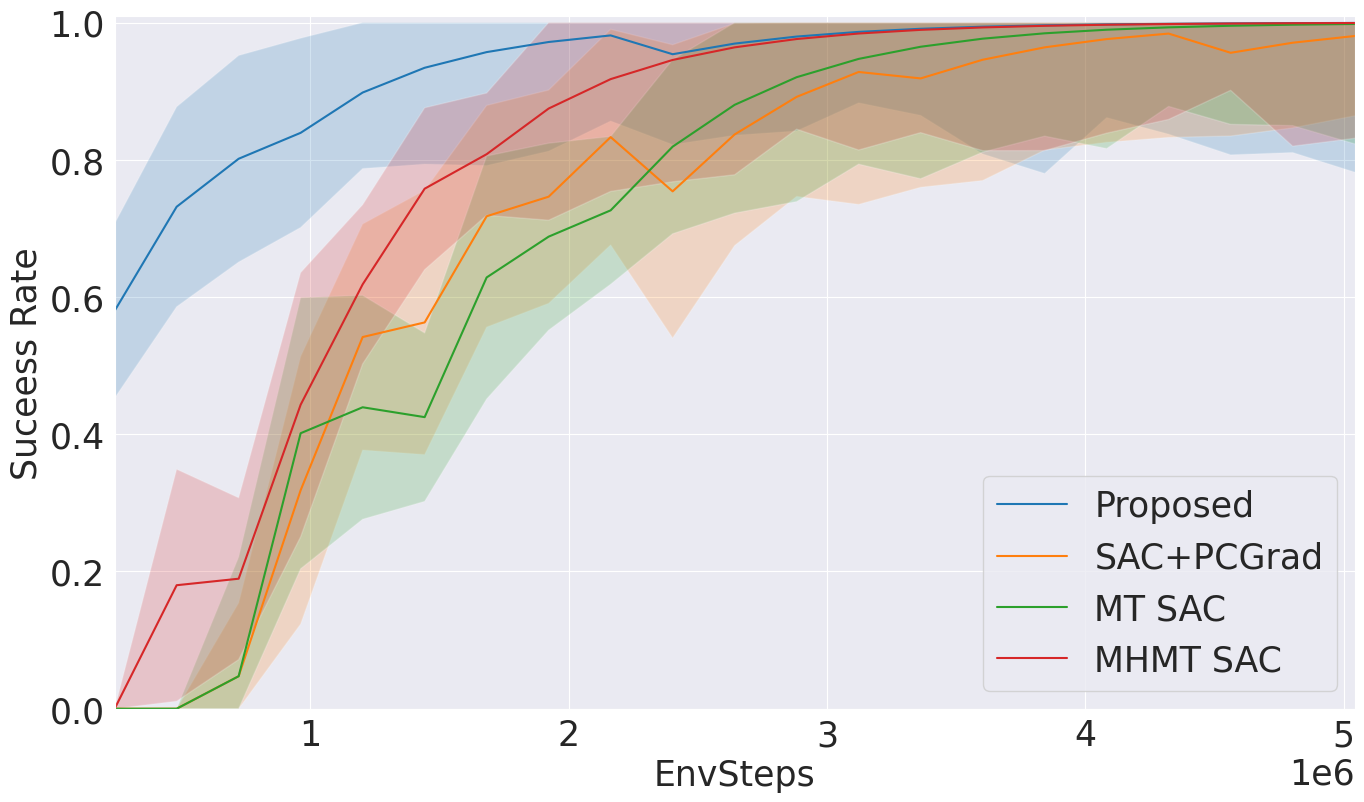}%
\vspace{-0.25cm}
\caption{Drawer Open}%
\label{Transfer to MTRL DrawerOpen:subfig:c}%
\end{subfigure}%
\medskip
\begin{subfigure}{0.25\textwidth}
\centering
\includegraphics[width=\linewidth, height=2.65cm]{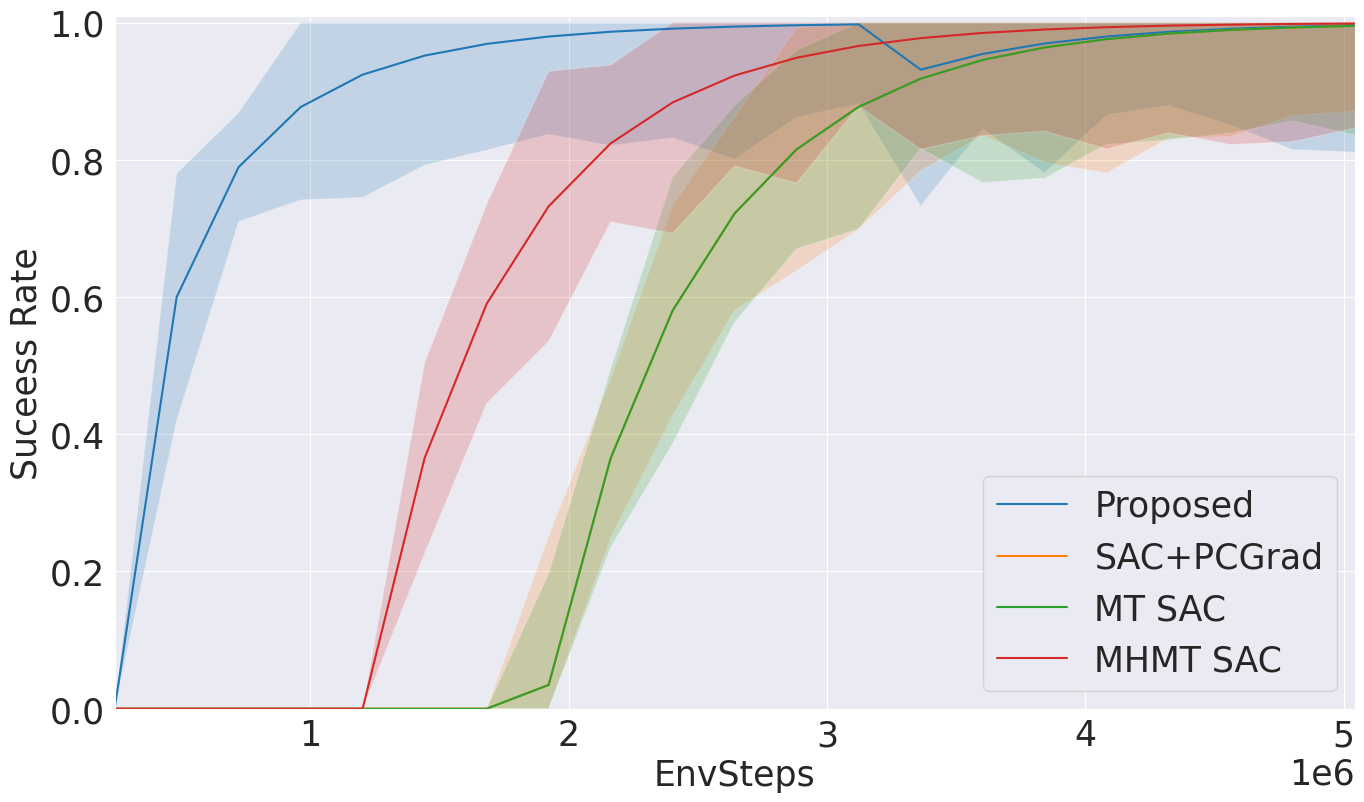}%
\vspace{-0.25cm}
\caption{Button Press}%
\label{Transfer to MTRL ButtonPress:subfig:d}%
\end{subfigure}%
\vspace{-0.25cm}

\begin{subfigure}{0.25\textwidth}
\centering
\includegraphics[width=\linewidth, height=2.65cm]{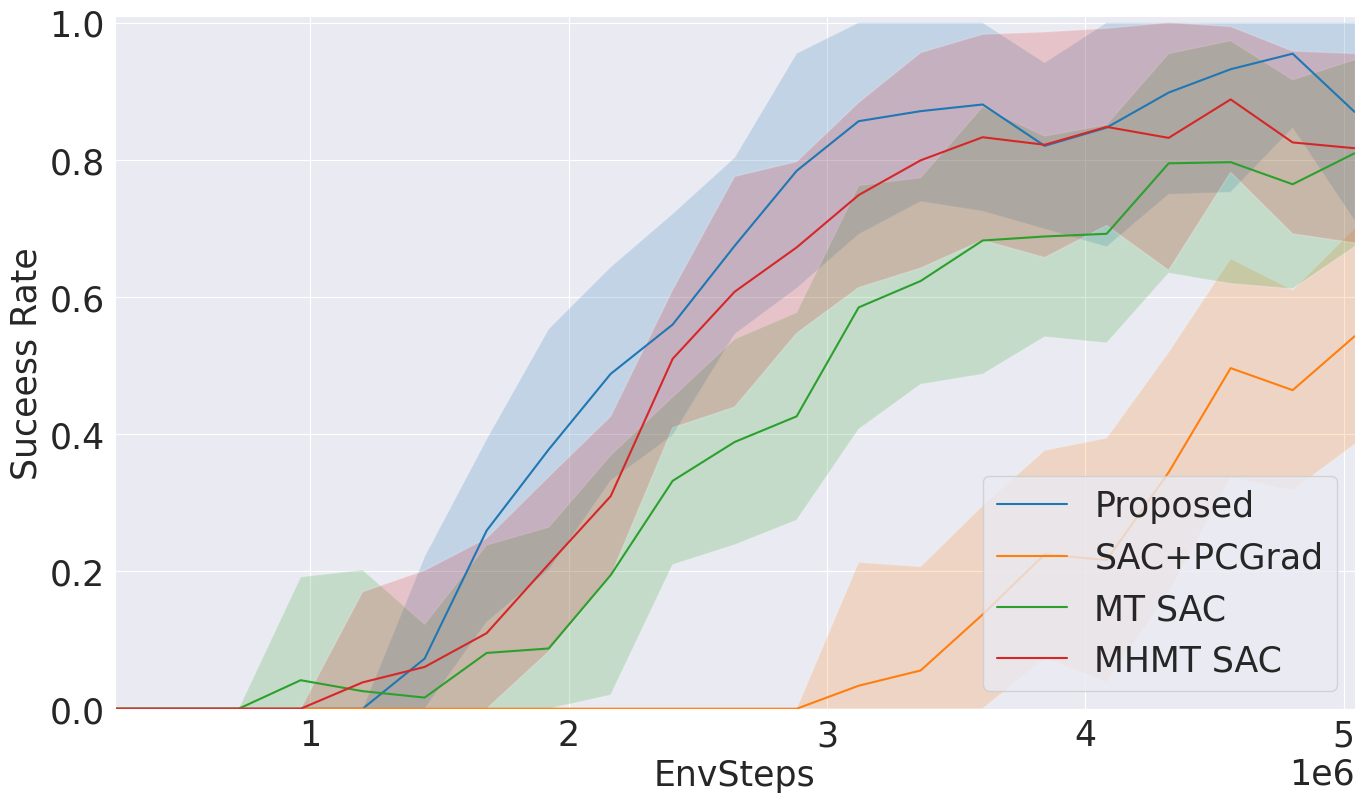}%
\vspace{-0.25cm}
\caption{Pushing}%
\label{Transfer to MTRL Pushing:subfig:e}%
\end{subfigure}%\vfill%
\medskip
\begin{subfigure}{0.25\textwidth}
\centering
\includegraphics[width=\linewidth, height = 2.65cm]{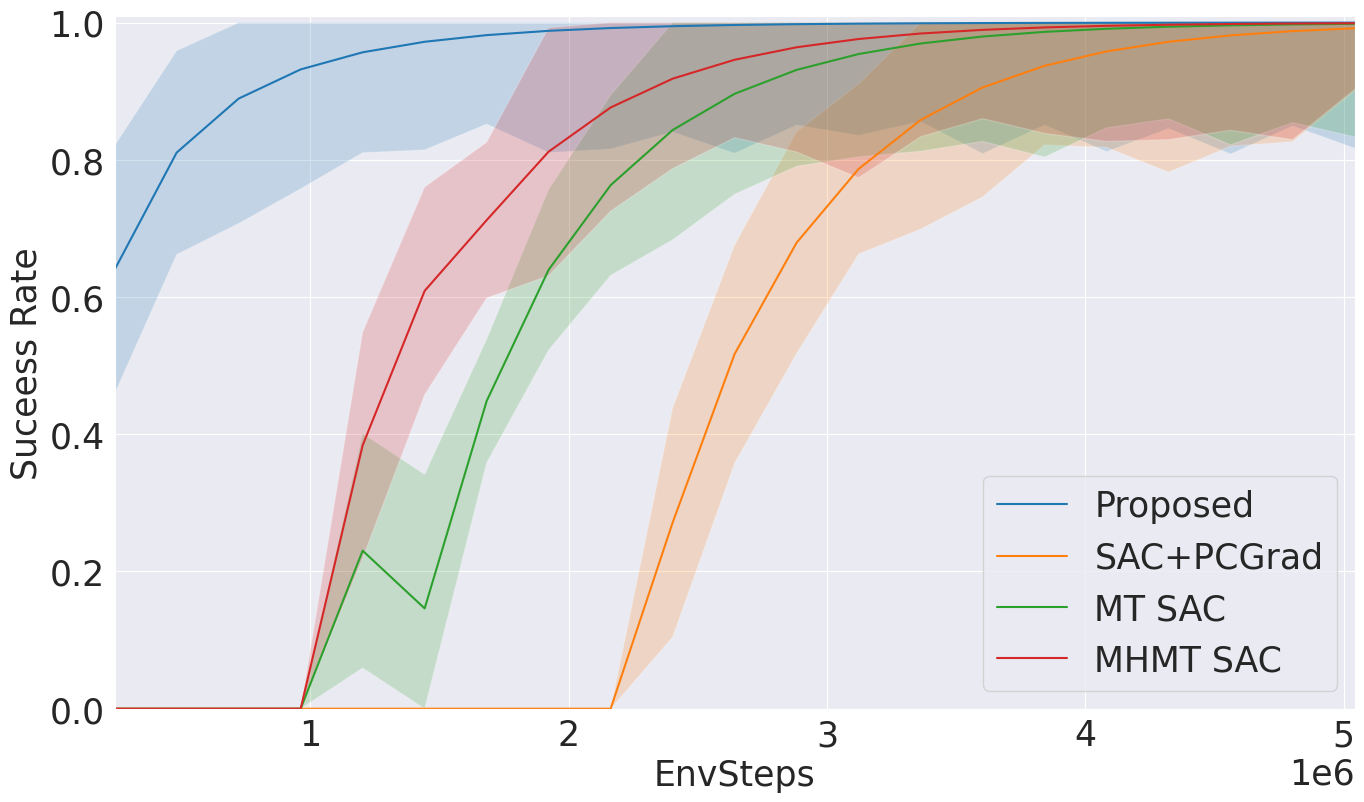}%
\vspace{-0.25cm}
\caption{Door Open}%
\label{Transfer to MTRL DoorOPen:subfig:f}%
\end{subfigure}%\hfill%
\medskip
\vspace{-0.5cm}
\caption{Some of the results in transferring to MTRL.} 
\label{fig:Transfer to MTRL}
\vspace{-0.6cm}
\end{figure}

As shown in Fig \ref{fig:Transfer to MTRL}, the proposed method achieves better sample efficiency and success rate in all of the tasks, even showing a higher initial success rate in a drawer, door environment due to the interactive property of the transferred primitives. Smaller performance gaps in the pushing task are due to the mismatch of action distribution as the gripper action is ignored to stay closed by default in this task. PCGrad is slower than other baselines, and it might be due to the assumption that there is a large enough conflict between tasks and large curvature in parameter space. But as we selectively choose the tasks that need interaction with an object and exclude the tasks such as reaching where no interaction is needed, there might be smaller gradient conflict than the metaworld environment. Other baselines MT-SAC and MHMT-SAC show better performance than PCGrad, but there are still large sample efficiency margins in the initial training process compared to the proposed one.

\section{Conclusion and Future Work}
In this work, we firstly present the unsupervised transferable manipulation skill discovery method for efficient adaptation to downstream tasks. We show that our method not only enables the agent to solve diverse, interaction-oriented exploration challenge without any external task reward but also distill the generated experience in the form of reusable task-agnostic skills. It brings significant improvement in sample efficiency when the skills are transferred to MTRL and GCRL even with multiple distractor objects. For future work, we would like to extend the idea into a reset-free or offline RL setting for developing more practical, real-world applicable algorithms.

\bibliographystyle{./bibtex/IEEEtran}
\bibliography{./bibtex/my_bib}

\begin{thebibliography}{10}
\providecommand{\url}[1]{#1}
\csname url@rmstyle\endcsname
\providecommand{\newblock}{\relax}
\providecommand{\bibinfo}[2]{#2}
\providecommand\BIBentrySTDinterwordspacing{\spaceskip=0pt\relax}
\providecommand\BIBentryALTinterwordstretchfactor{4}
\providecommand\BIBentryALTinterwordspacing{\spaceskip=\fontdimen2\font plus
\BIBentryALTinterwordstretchfactor\fontdimen3\font minus
  \fontdimen4\font\relax}
\providecommand\BIBforeignlanguage[2]{{%
\expandafter\ifx\csname l@#1\endcsname\relax
\typeout{** WARNING: IEEEtran.bst: No hyphenation pattern has been}%
\typeout{** loaded for the language `#1'. Using the pattern for}%
\typeout{** the default language instead.}%
\else
\language=\csname l@#1\endcsname
\fi
#2}}

\bibitem{gu2017deep}
S.~Gu, E.~Holly, T.~Lillicrap, and S.~Levine, ``Deep reinforcement learning for
  robotic manipulation with asynchronous off-policy updates,'' in \emph{2017
  IEEE international conference on robotics and automation (ICRA)}.\hskip 1em
  plus 0.5em minus 0.4em\relax IEEE, 2017, pp. 3389--3396.

\bibitem{yamada2020motion}
J.~Yamada, Y.~Lee, G.~Salhotra, K.~Pertsch, M.~Pflueger, G.~S. Sukhatme, J.~J.
  Lim, and P.~Englert, ``Motion planner augmented reinforcement learning for
  robot manipulation in obstructed environments,'' \emph{arXiv preprint
  arXiv:2010.11940}, 2020.

\bibitem{sharma2020emergent}
A.~Sharma, M.~Ahn, S.~Levine, V.~Kumar, K.~Hausman, and S.~Gu, ``Emergent
  real-world robotic skills via unsupervised off-policy reinforcement
  learning,'' \emph{arXiv preprint arXiv:2004.12974}, 2020.

\bibitem{eysenbach2018diversity}
B.~Eysenbach, A.~Gupta, J.~Ibarz, and S.~Levine, ``Diversity is all you need:
  Learning skills without a reward function,'' \emph{arXiv preprint
  arXiv:1802.06070}, 2018.

\bibitem{pathak2017curiosity}
D.~Pathak, P.~Agrawal, A.~A. Efros, and T.~Darrell, ``Curiosity-driven
  exploration by self-supervised prediction,'' in \emph{International
  conference on machine learning}.\hskip 1em plus 0.5em minus 0.4em\relax PMLR,
  2017, pp. 2778--2787.

\bibitem{burda2018exploration}
Y.~Burda, H.~Edwards, A.~Storkey, and O.~Klimov, ``Exploration by random
  network distillation,'' \emph{arXiv preprint arXiv:1810.12894}, 2018.

\bibitem{bellemare2016unifying}
M.~Bellemare, S.~Srinivasan, G.~Ostrovski, T.~Schaul, D.~Saxton, and R.~Munos,
  ``Unifying count-based exploration and intrinsic motivation,'' \emph{Advances
  in neural information processing systems}, vol.~29, 2016.

\bibitem{ostrovski2017count}
G.~Ostrovski, M.~G. Bellemare, A.~Oord, and R.~Munos, ``Count-based exploration
  with neural density models,'' in \emph{International conference on machine
  learning}.\hskip 1em plus 0.5em minus 0.4em\relax PMLR, 2017, pp. 2721--2730.

\bibitem{machado2017laplacian}
M.~C. Machado, M.~G. Bellemare, and M.~Bowling, ``A laplacian framework for
  option discovery in reinforcement learning,'' in \emph{International
  Conference on Machine Learning}.\hskip 1em plus 0.5em minus 0.4em\relax PMLR,
  2017, pp. 2295--2304.

\bibitem{warde2018unsupervised}
D.~Warde-Farley, T.~Van~de Wiele, T.~Kulkarni, C.~Ionescu, S.~Hansen, and
  V.~Mnih, ``Unsupervised control through non-parametric discriminative
  rewards,'' \emph{arXiv preprint arXiv:1811.11359}, 2018.

\bibitem{pathak2019self}
D.~Pathak, D.~Gandhi, and A.~Gupta, ``Self-supervised exploration via
  disagreement,'' in \emph{International conference on machine learning}.\hskip
  1em plus 0.5em minus 0.4em\relax PMLR, 2019, pp. 5062--5071.

\bibitem{sharma2019dynamics}
A.~Sharma, S.~Gu, S.~Levine, V.~Kumar, and K.~Hausman, ``Dynamics-aware
  unsupervised discovery of skills,'' \emph{arXiv preprint arXiv:1907.01657},
  2019.

\bibitem{zhao2020mutual}
R.~Zhao, Y.~Gao, P.~Abbeel, V.~Tresp, and W.~Xu, ``Mutual information-based
  state-control for intrinsically motivated reinforcement learning,''
  \emph{arXiv preprint arXiv:2002.01963}, 2020.

\bibitem{devin2017learning}
C.~Devin, A.~Gupta, T.~Darrell, P.~Abbeel, and S.~Levine, ``Learning modular
  neural network policies for multi-task and multi-robot transfer,'' in
  \emph{2017 IEEE international conference on robotics and automation
  (ICRA)}.\hskip 1em plus 0.5em minus 0.4em\relax IEEE, 2017, pp. 2169--2176.

\bibitem{qureshi2019composing}
A.~H. Qureshi, J.~J. Johnson, Y.~Qin, T.~Henderson, B.~Boots, and M.~C. Yip,
  ``Composing task-agnostic policies with deep reinforcement learning,''
  \emph{arXiv preprint arXiv:1905.10681}, 2019.

\bibitem{yu2020meta}
T.~Yu, D.~Quillen, Z.~He, R.~Julian, K.~Hausman, C.~Finn, and S.~Levine,
  ``Meta-world: A benchmark and evaluation for multi-task and meta
  reinforcement learning,'' in \emph{Conference on Robot Learning}.\hskip 1em
  plus 0.5em minus 0.4em\relax PMLR, 2020, pp. 1094--1100.

\bibitem{yang2020multi}
R.~Yang, H.~Xu, Y.~Wu, and X.~Wang, ``Multi-task reinforcement learning with
  soft modularization,'' \emph{arXiv preprint arXiv:2003.13661}, 2020.

\bibitem{yu2020gradient}
T.~Yu, S.~Kumar, A.~Gupta, S.~Levine, K.~Hausman, and C.~Finn, ``Gradient
  surgery for multi-task learning,'' \emph{arXiv preprint arXiv:2001.06782},
  2020.

\bibitem{suteu2019regularizing}
M.~Suteu and Y.~Guo, ``Regularizing deep multi-task networks using orthogonal
  gradients,'' \emph{arXiv preprint arXiv:1912.06844}, 2019.

\bibitem{peng2019mcp}
X.~B. Peng, M.~Chang, G.~Zhang, P.~Abbeel, and S.~Levine, ``Mcp: Learning
  composable hierarchical control with multiplicative compositional policies,''
  in \emph{Advances in Neural Information Processing Systems}, 2019, pp.
  3686--3697.

\bibitem{nowozin2016f}
S.~Nowozin, B.~Cseke, and R.~Tomioka, ``f-gan: Training generative neural
  samplers using variational divergence minimization,'' in \emph{Proceedings of
  the 30th International Conference on Neural Information Processing Systems},
  2016, pp. 271--279.

\bibitem{haarnoja2018soft}
T.~Haarnoja, A.~Zhou, P.~Abbeel, and S.~Levine, ``Soft actor-critic: Off-policy
  maximum entropy deep reinforcement learning with a stochastic actor,'' in
  \emph{International conference on machine learning}.\hskip 1em plus 0.5em
  minus 0.4em\relax PMLR, 2018, pp. 1861--1870.

\bibitem{vaswani2017attention}
A.~Vaswani, N.~Shazeer, N.~Parmar, J.~Uszkoreit, L.~Jones, A.~N. Gomez,
  {\L}.~Kaiser, and I.~Polosukhin, ``Attention is all you need,'' in
  \emph{Advances in neural information processing systems}, 2017, pp.
  5998--6008.

\bibitem{parisotto2020stabilizing}
E.~Parisotto, F.~Song, J.~Rae, R.~Pascanu, C.~Gulcehre, S.~Jayakumar,
  M.~Jaderberg, R.~L. Kaufman, A.~Clark, S.~Noury, \emph{et~al.}, ``Stabilizing
  transformers for reinforcement learning,'' in \emph{International Conference
  on Machine Learning}.\hskip 1em plus 0.5em minus 0.4em\relax PMLR, 2020, pp.
  7487--7498.

\bibitem{tsai2020neural}
Y.-H.~H. Tsai, H.~Zhao, M.~Yamada, L.-P. Morency, and R.~Salakhutdinov,
  ``Neural methods for point-wise dependency estimation,'' \emph{arXiv preprint
  arXiv:2006.05553}, 2020.

\bibitem{tsai2021self}
Y.-H.~H. Tsai, M.~Q. Ma, M.~Yang, H.~Zhao, L.-P. Morency, and R.~Salakhutdinov,
  ``Self-supervised representation learning with relative predictive coding,''
  \emph{arXiv preprint arXiv:2103.11275}, 2021.

\bibitem{rusu2016progressive}
A.~A. Rusu, N.~C. Rabinowitz, G.~Desjardins, H.~Soyer, J.~Kirkpatrick,
  K.~Kavukcuoglu, R.~Pascanu, and R.~Hadsell, ``Progressive neural networks,''
  \emph{arXiv preprint arXiv:1606.04671}, 2016.

\bibitem{andrychowicz2017hindsight}
M.~Andrychowicz, F.~Wolski, A.~Ray, J.~Schneider, R.~Fong, P.~Welinder,
  B.~McGrew, J.~Tobin, P.~Abbeel, and W.~Zaremba, ``Hindsight experience
  replay,'' \emph{arXiv preprint arXiv:1707.01495}, 2017.

\end{thebibliography}

\addtolength{\textheight}{-12cm}   % This command serves to balance the column lengths

%%%%%%%%%%%%%%%%%%%%%%%%%%%%%%%%%%%%%%%%%%%%%%%%%%%%%%%%%%%%%%%%%%%%%%%%%%%%%%%%

%%%%%%%%%%%%%%%%%%%%%%%%%%%%%%%%%%%%%%%%%%%%%%%%%%%%%%%%%%%%%%%%%%%%%%%%%%%%%%%%



% \section*{ACKNOWLEDGMENT}

% The preferred spelling of the word ÒacknowledgmentÓ in America is without an ÒeÓ after the ÒgÓ. Avoid the stilted expression, ÒOne of us (R. B. G.) thanks . . .Ó  Instead, try ÒR. B. G. thanksÓ. Put sponsor acknowledgments in the unnumbered footnote on the first page.

%%%%%%%%%%%%%%%%%%%%%%%%%%%%%%%%%%%%%%%%%%%%%%%%%%%%%%%%%%%%%%%%%%%%%%%%%%%%%%%%

% References are important to the reader; therefore, each citation must be complete and correct. If at all possible, references should be commonly available publications.

% \begin{thebibliography}{99}
% \bibitem{c1} G. O. Young, ÒSynthetic structure of industrial plastics (Book style with paper title and editor),Ó 	in Plastics, 2nd ed. vol. 3, J. Peters, Ed.  New York: McGraw-Hill, 1964, pp. 15Ð64.
% \end{thebibliography}

\end{document}